\documentclass[lettersize,journal]{IEEEtran}
\usepackage{amsmath,amsfonts}
\usepackage{algorithmic}
\usepackage{algorithm}
\usepackage{array}
\usepackage[caption=false,font=normalsize,labelfont=sf,textfont=sf]{subfig}
\usepackage{textcomp}
\usepackage{stfloats}
\usepackage{url}
\usepackage{verbatim}
\usepackage{graphicx}
\usepackage{cite,color}
\usepackage{multicol}
\usepackage{multirow}
\usepackage{makecell}
\usepackage{booktabs}
\begin{document}


\title{Generative Artificial Intelligence Meets Synthetic Aperture Radar: A Survey}


\author{Zhongling Huang,~\IEEEmembership{Member,~IEEE,}
Xidan Zhang,
Zuqian Tang,
Feng Xu,~\IEEEmembership{Senior Member,~IEEE,}\\
Mihai Datcu,~\IEEEmembership{Fellow,~IEEE,}
Junwei Han,~\IEEEmembership{Fellow,~IEEE}
\thanks{This work was supported by the National Natural Science Foundation of China under Grant 62101459.}
}

\markboth{Accepted by IEEE Geoscience and Remote Sensing Magazine}%
{Shell \MakeLowercase{\textit{et al.}}: A Sample Article Using IEEEtran.cls for IEEE Journals}


\maketitle

\begin{figure*}[!hbp]
    \centering
    \includegraphics[width=0.55\textwidth]{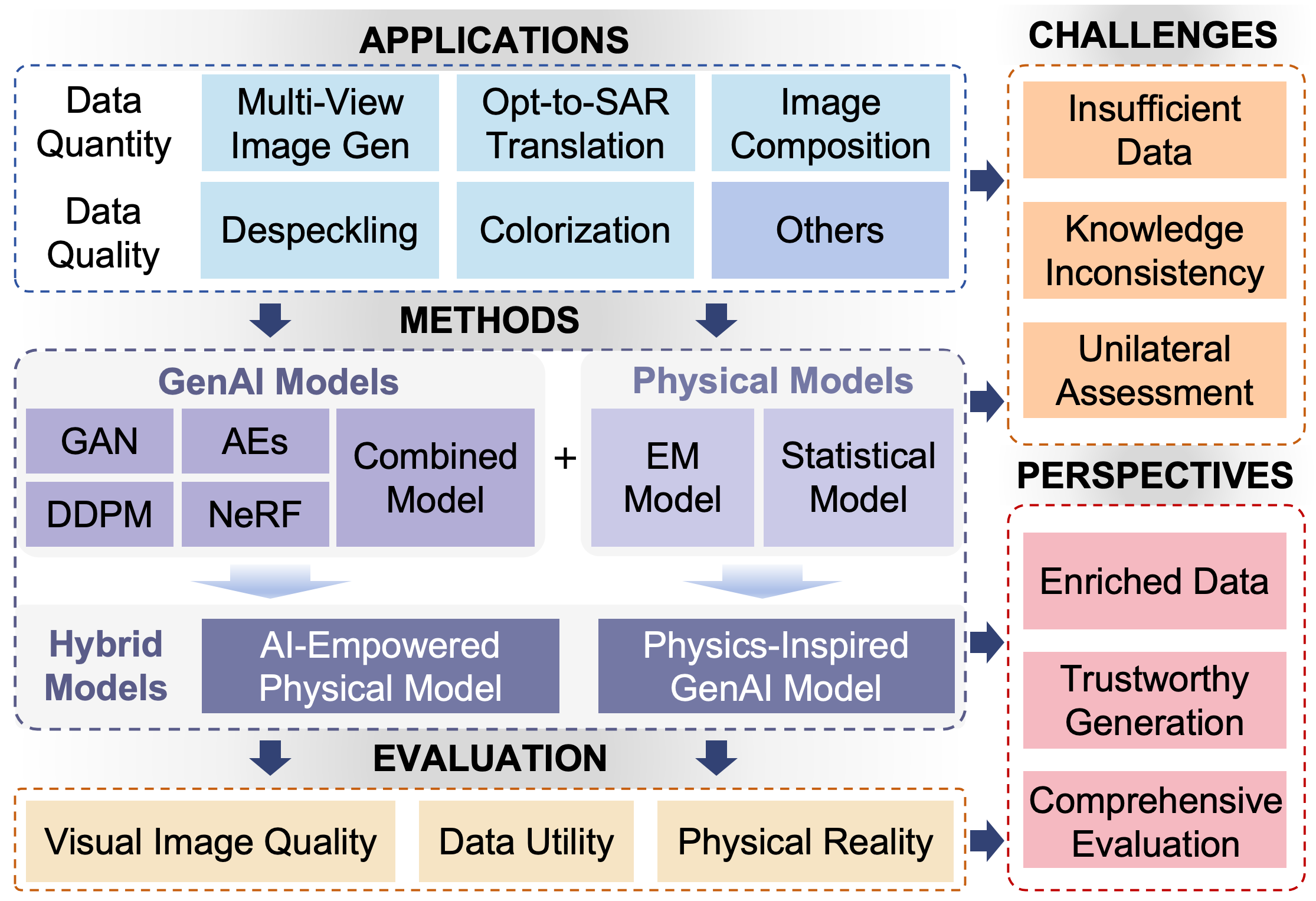}
    \caption{The outline of this survey.}
    \label{fig:intro}
\end{figure*}

\begin{abstract}
SAR images possess unique attributes that present challenges for both human observers and vision AI models to interpret, owing to their electromagnetic characteristics. The interpretation of SAR images encounters various hurdles, with one of the primary obstacles being the data itself, which includes issues related to both the quantity and quality of the data. The challenges can be addressed using generative AI technologies. Generative AI, often known as GenAI, is a very advanced and powerful technology in the field of artificial intelligence that has gained significant attention. The advancement has created possibilities for the creation of texts, photorealistic pictures, videos, and material in various modalities. This paper aims to comprehensively investigate the intersection of GenAI and SAR. First, we illustrate the common data generation-based applications in SAR field and compare them with computer vision tasks, analyzing the similarity, difference, and general challenges of them. Then, an overview of the latest GenAI models is systematically reviewed, including various basic models and their variations targeting the general challenges. Additionally, the corresponding applications in SAR domain are also included. Specifically, we propose to summarize the physical model based simulation approaches for SAR, and analyze the hybrid modeling methods that combine the GenAI and interpretable models. The evaluation methods that have been or could be applied to SAR, are also explored. Finally, the potential challenges and future prospects are discussed. To our best knowledge, this survey is the first exhaustive examination of the interdiscipline of SAR and GenAI, encompassing a wide range of topics, including deep neural networks, physical models, computer vision, and SAR images. The resources of this survey are open-source at \url{https://github.com/XAI4SAR/GenAIxSAR}.
\end{abstract}


\section{Motivation and Significance of The Topic}



Synthetic Aperture Radar (SAR) has garnered significant interest in a wide range of earth observation applications \cite{tuia2023artificial,chenNonparametricFullApertureAutofocus2024,chenFullApertureProcessingAirborne2023,gaoOnboardInformationFusion2023,zhangDevelopmentApplicationShip2024}. SAR images possess unique attributes that present challenges for both human observers and vision AI models to interpret, owing to their electromagnetic characteristics \cite{huangPhysicallyExplainableCNN2022,huangClassificationLargeScaleHighResolution2021,huangHDECTFAUnsupervisedLearning2021,huangDeepSARNetLearning2020}. The analysis of SAR images has numerous challenges, with one of the main hindrances being the data itself, encompassing concerns regarding both the volume and the quality of the data. The backscattering properties of objects display notable fluctuations depending on elements such as the frequency of operation, polarization, and viewing angles. As a result, this creates significant inconsistencies in the data and restricts the generalization ability of an AI model trained on that data. There is a pressing requirement to generate additional data in different application scenarios in order to meet the criterion for training data. Moreover, it is frequently important to augment the ability to understand SAR images by enhancing the quality of the data. The presence of sensor-specific speckle noise and azimuth ambiguity, for example, significantly impairs the image quality, making it challenging to comprehend \cite{huangUncertaintyExplorationExplainable2023}.

The aforementioned challenges can be addressed using Generative AI technologies. Generative AI, often known as GenAI, is a very advanced and powerful technology in the field of artificial intelligence that has gained significant attention. The field has shown significant growth since the start of generative adversarial networks (GANs) in 2014, and particularly accelerated with the emergence of multi-modality foundation models recently. With the emergence of various powerful GenAI models, such as GPT (OpenAI) \cite{radford2018improving}, LaMDA (Google) \cite{thoppilan2022lamda}, LLaMA (Meta) \cite{touvron2023llama} for text generation, DALL$\cdot$E 3 \cite{betker2023improving}, Stable Diffusion \cite{rombach2022high}, Imagen \cite{saharia2022photorealistic}, and Midjourney \cite{Midjourney2022} for text-to-image generation, Sora \cite{liu2024sora} for text-to-video generation, the field has entered a new age. The advancement has created possibilities for the creation of texts, photorealistic pictures, videos, and material in various modalities. The created information emulates human language and visual perception, facilitating human comprehension of the world \cite{zhanMultimodalImageSynthesis2023}. The fundamental technologies behind them consist of Transformer-based large language models (LLMs), vision foundation models (VFMs), GANs, Diffusion models, etc. For instance, the LLMs are employed by the text-to-image GenAI models to encode the text prompt as semantic embedding. Subsequently, it is employed to condition the generator in order to generate the image. This can be accomplished through the application of diffusion models, as demonstrated by Imagen and DALL$\cdot$E 3.

\begin{figure*}[!hbp]
    \centering
    \includegraphics[width=0.9\textwidth]{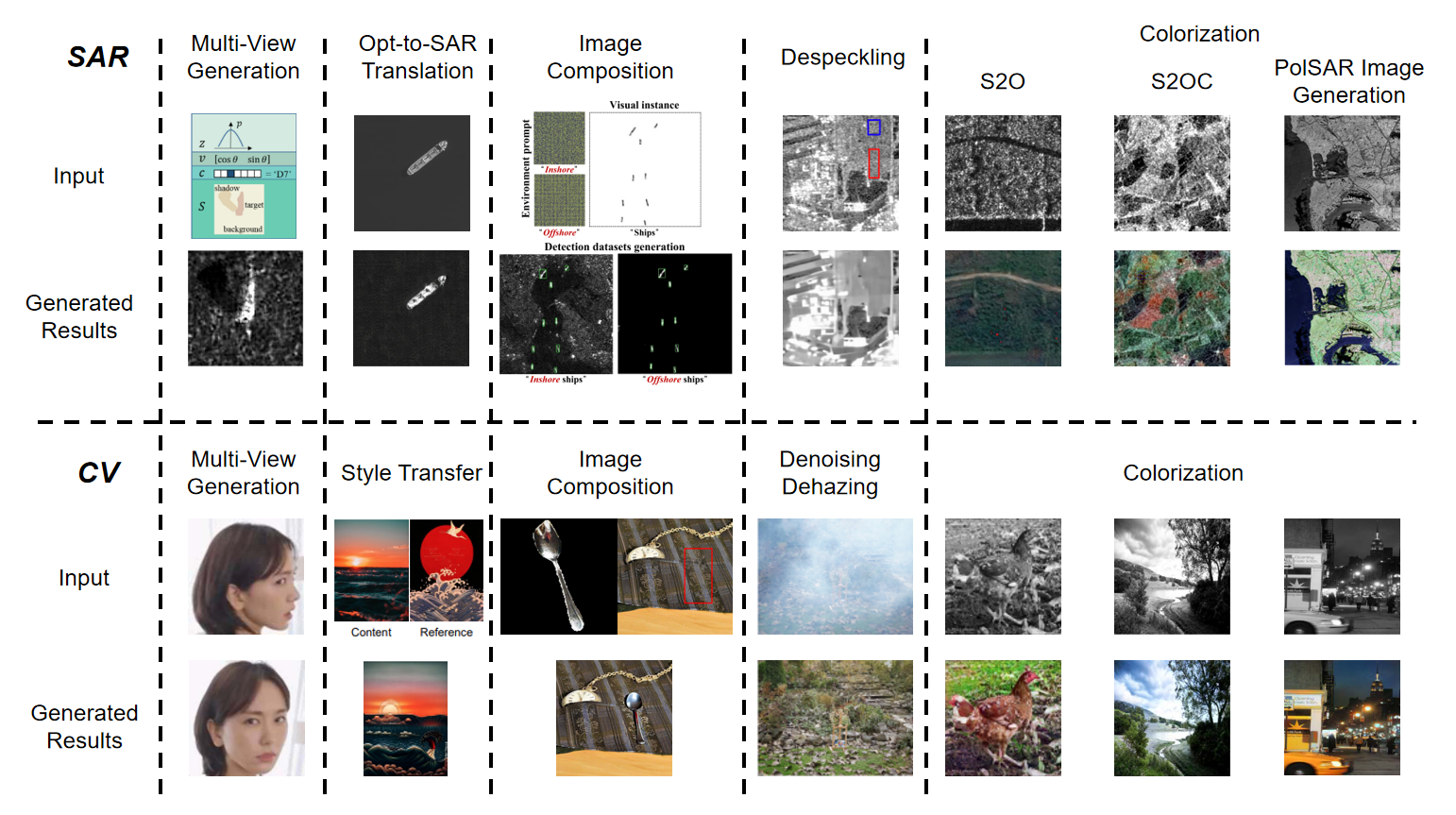}
    \caption{Illustrations of different applications in SAR and computer vision domain. The samples are from \cite{songLearningGenerateSAR2022,shiUnsupervisedDomainAdaptation2022,zhang2024ship,pereraSARDespecklingUsing2023,wangSARtoOpticalImageTranslation2022,shenBenchmarkingProtocolSAR2024,songRadarImageColorization2018,xuMultiViewFaceSynthesis2021,zhangInversionbasedStyleTransfer2023,songObjectStitchObjectCompositing2023,fuDWGANDiscreteWavelet2021,wuVividDiverseImage2021}.}
    \label{fig:task}
\end{figure*}

In the field of computer vision, the problem of image generation and reconstruction via GenAI models has been thoroughly discussed and explored. There are, however, some important differences when GenAI and SAR collide. Most of the existing GenAI models exhibit a deficiency in SAR knowledge and a lack of awareness regarding the intricate physical properties and complex nature of SAR data. In pursuit of this objective, they encounter limitations in their ability to produce a wide range of SAR images that possess appropriate physical characteristics, including polarimetry and interferometry. For SAR data, the vision-language model, often known as the vision GenAI model, would not be effective. In addition, the process of generating images using Synthetic Aperture Radar (SAR) should take into account certain physical properties, such as sub-aperture coherency. Moreover, the assessment of generated synthetic aperture radar (SAR) images may vary from that of optical data due to the inherent differences in their imaging mechanisms, thereby rendering it unreliable. It is of utmost importance to highlight the particulars of problem definition and the potential challenges associated with it in order to effectively identify multiple solutions for SAR data.


The extant literature reviews have furnished a clear overview of the applications of auto-encoders and GAN-based models in the realm of SAR/ISAR, specifically in the domain of despeckling \cite{zhangApplicationDeepGenerative2023,fracastoroDeepLearningMethods2021}. Nevertheless, a number of cutting-edge GenAI models were excluded, and there is a scarcity of comprehensive comparisons and analyses with other pertinent domains that could provide valuable insights for SAR applications. More specifically, the survey is conducted by examining various aspects, including applications, methods, evaluations, and perspectives. We not only provide a summary of the current approaches for generating SAR images using deep learning, but also conduct a comprehensive analysis of the literature in the related domain that addresses the general challenges associated with SAR. The outline of this survey is briefly illustrated in Fig. \ref{fig:intro}. 

Despite some initial accomplishments, the full potential of GenAI in the field of SAR remains largely unexplored. Illustrative instances encompass the utilization of 3D Gaussian splatting and neural radiance fields for the purpose of three-dimensional reconstruction and rendering of SAR targets, as well as the simulation of expansive SAR scene images achieved through the text-to-image foundation GenAI model. There are undoubtedly numerous challenges associated with the transfer of advanced artificial intelligence (AI) technologies from the computer vision domain to SAR applications. The objective of this paper is to promote further research in this captivating yet underexplored field of study.

\section{Applications}

The limitations in both the quality and quantity of data are significant obstacles that impede the progress of deep learning-based SAR image interpretation, as previously discussed. In order to accomplish this objective, the primary aims of utilising GenAI models in the context of SAR involve the augmentation of images and the enhancement of image quality. Although computer vision applications are prevalent, there are significant differences in the case of SAR. This section will comprehensively include the comparisons. The illustrations are provided in Fig. \ref{fig:task} to demonstrate the typical applications with similarity and difference in SAR and vision domain. Table \ref{tab:applications} enumerates the representative studies of different applications, and highlights the specific characteristics of SAR.

\begin{table*}[h!]
    \renewcommand{\arraystretch}{1.5}
    \centering
    \caption{The tasks of image generation and reconstruction have some significant difference for SAR and vision applications.}
    \label{tab:applications}
    \begin{tabular}{cccccc}
    \toprule
     &\multicolumn{2}{c}{\textbf{Task}} &\textbf{SAR} &\textbf{Vision} &\textbf{Specific Characteristics of SAR}\\
    \midrule
    \multirow{3}{*}{\makecell[c]{\textbf{Towards} \\ \textbf{Increasing} \\ \textbf{Data} \\ \textbf{Quantity}}}
    &\multicolumn{2}{c}{\makecell[c]{Multi-View \\ Image Generation}}
    &\cite{wangSARTargetImage2022,daiCVGANCrossViewSAR2023,giry2022sar,shiISAGANHighFidelityFullAzimuth2022,sunAttributeGuidedGenerativeAdversarial2023,ohPeaceGANGANBasedMultiTask2021,huFeatureLearningSAR2021,songSARImageRepresentation2019,songLearningGenerateSAR2022,guoCausalAdversarialAutoencoder2023,lei2024sar}
    &\makecell[c]{\cite{xuMultiViewFaceSynthesis2021,chenMultiViewGaitImage2021,yin2020novel}} & \makecell[c]{Electromagnetic features \\ Imaging geometry \\ Distortion effect}\\
    \cline{2-6}
    
    &\multicolumn{2}{c}{\makecell[c]{O2S}}
    &\cite{shiUnsupervisedDomainAdaptation2022,rangzan2023tsgan,fuReciprocalTranslationSAR2021}
    &\makecell[c]{(Style Transfer) \\ \cite{zhangInversionbasedStyleTransfer2023,chenCartoonGANGenerativeAdversarial2018}} & \makecell[c]{Ill-posed task \\ Sensor-specific solution} \\
    \cline{2-6}

    &\multicolumn{2}{c}{\makecell[c]{Image Composition}} &\cite{zhang2024ship,sunDSDetLightweightDensely2021,kuangSemanticLayoutGuidedImageSynthesis2023,luanBCNetBackgroundConversion2024}
    &\cite{songObjectStitchObjectCompositing2023,zhang2020learning} & \makecell[c]{Consistent geometry distortion \\ for foreground and background} \\
    
    
    \hline
    \multirow{11}{*}{\makecell[c]{\textbf{Toward} \\ \textbf{Improving} \\ \textbf{Data} \\ \textbf{Quality}}}
    &\multicolumn{2}{c}{Despeckling}
    &\cite{pereraSARDespecklingUsing2023,lattariCycleSARSARImage2023,wangSARImageDespeckling2022,koSARImageDespeckling2022,hu2024sar} & \makecell[c]{(Denoising) \\ \cite{chenImageBlindDenoising2018,fuDWGANDiscreteWavelet2021}} & \makecell[c]{Different statistical models \\ for speckle noise} \\
    \cline{2-6}
    
    &\multirow{3}{*}{Colorization}
    &S2O
    &\makecell[c]{\cite{baiConditionalDiffusionSAR2024,shiBraininspiredApproachSARtooptical2024,zhangComparativeAnalysisEdge2021,yangSARtoopticalImageTranslation2022,niuImageTranslationHighresolution2021,leeCFCASETCoarsetoFineContextAware2023,zhangSARtoOpticalImageTranslation2024,weiCFRWDGANSARtoOpticalImage2023}\\
    \cite{wangSARtoOpticalImageTranslation2022,liDeepTranslationGAN2021,fuReciprocalTranslationSAR2021,liMultiscaleGenerativeAdversarial2022,guoEdgePreservingConvolutionalGenerative2021,yangFGGANFineGrainedGenerative2022,wangGeneratingHighQuality2018}}
    &\multirow{3}{*}{\makecell[c]{\cite{nazeriImageColorizationUsing2018,wuVividDiverseImage2021}}} & \multirow{3}{*}{\makecell[c]{S2O: multi-modal translation \\ S2Oc: preserving texture of SAR \\ PolGen: reconstruct the polarimetric \\ features}}\\
    \cline{3-4}
    & &S2Oc
    &\cite{schmitt2018colorizing,shenBenchmarkingProtocolSAR2024,wangGeneratingHighQuality2018} & &\\
    \cline{3-4}
    & &\makecell[c]{PolSAR Generation}
    &\cite{songRadarImageColorization2018} & &\\
    
    \cline{2-6}
     &\multirow{5}{*}{Others}
    & \makecell[c]{Super \\ Resolution} &\cite{guSARImageSuperResolution2019, shenResidualConvolutionalNeural2020,heLearningBasedCompressed2012,kongDMSCGANCGANBasedFramework2023}
    &\cite{zhangSupervisedPixelWiseGAN2021,liSRDiffSingleImage2022}
    & Different degradation model \\
    \cline{3-6}
    & &\makecell[c]{Layout Scene \\ Image Generation} &\cite{juSARGANNovelSAR2023} 
    &\cite{zhaoImageGenerationLayout2019}
    & SAR image features \\
    \cline{3-6}
    & & Interference Suppression & \cite{WEI2022103019,oyedareInterferenceSuppressionUsing2022,zhangInterferenceSuppressionAlgorithm2011} 
    &\makecell[c]{(Cloud removal / dehazing) \\ \cite{fuDWGANDiscreteWavelet2021}}
    & Electromagnetic interference \\
    \bottomrule
    \end{tabular}
\end{table*}

\subsection{Towards Increasing Data Quantity}

The limited availability of training data has impeded the advancement of deep learning methods in the domain of SAR image interpretation. The importance lies not in the amount of SAR data, but rather in the abundance of SAR images in specific application scenarios \cite{ghozatlouGanBasedOceanPattern2023}. This section introduces three common applications: multi-view SAR image synthesis, optical-to-SAR (O2S) translation, and SAR image composition.

\subsubsection{Multi-View Image Generation}

The community is primarily focused on the issue of generating multi-view SAR images, which has significant practical implications for SAR target detection and recognition when there is a scarcity of training data. This task has been highlighted in the literature by Zhang et al. \cite{zhangApplicationDeepGenerative2023}. The objective is to generate new images from alternative perspectives using a collection of images captured from restricted angles. The multi-view image synthesis or 3D reconstruction work shows a substantial difference due to the unique imaging mechanisms of optical and SAR. Optical images are captured by optical sensors, such as cameras, and rely on light within the visible and near-infrared spectral ranges. They capture optical features such as the color, shape, and texture of the target surface. The objective of multi-view image generation for optical images is to recreate the optical characteristics from several perspectives, simulating the effect of being taken by multiple cameras. As a comparison, SAR system emits a pulsed electromagnetic wave, which interacts with the target surface upon reflection. The radar antenna captures the backscattered signal, which contains information about the target’s properties. The electromagnetic features such as backscattering intensity, multiple bouncing effects, instead of optical features of color and shape, are required in multi-view SAR image generation. Moreover, the scattering properties of SAR targets exhibit considerable variation depending on imaging factors like aspect angle and sensor resolution. This poses a greater challenge in generating many views with limited observations. Many researches focus on this task including but not limit to \cite{wangSARTargetImage2022,daiCVGANCrossViewSAR2023,giry2022sar,shiISAGANHighFidelityFullAzimuth2022,sunAttributeGuidedGenerativeAdversarial2023,ohPeaceGANGANBasedMultiTask2021,huFeatureLearningSAR2021,songSARImageRepresentation2019,songLearningGenerateSAR2022,guoCausalAdversarialAutoencoder2023,lei2024sar} and we will illustrate them elaborately in the later sections.

\subsubsection{Optical-to-SAR Translation}

The goal of the optical-to-SAR (O2S) translation task is to utilize optical time series data to complete the missing information in the SAR time series data, either to augment SAR data or to improve SAR's spatial resolution. In contrast to the well-researched field of SAR-to-Optical translation, O2S is more difficult to tackle because of the ill-posed nature of the translation. The complexity occurs due to the fact that a same optical data might have various SAR representations, depending on the SAR viewing geometry.

Several studies have concentrated on this particular task, including \cite{shiUnsupervisedDomainAdaptation2022,rangzan2023tsgan,fuReciprocalTranslationSAR2021}. Fu et al. \cite{fuReciprocalTranslationSAR2021} introduced a modified Pix2Pix architecture with multiscale cascaded residual connections for SAR-optical reciprocal translation. The pixel-domain adaptation (PDA) proposed by Shi \cite{shiUnsupervisedDomainAdaptation2022} employed CycleGAN to generate a transition domain. The transition domain and the source domain was then combined to enrich the insufficient SAR data for training and improve the detection model's capacity to generalize. The proposed Temporal Shifting GAN (TSGAN) \cite{rangzan2023tsgan} is a dual conditional GAN designed for optical-to-SAR translation. The model was built based on a novel attention-based siamese encoder, and a change weighted loss function was also employed to prevent the TSGAN from overfitting. The TSGAN obtained foundational geometry from the prior timestamp SAR data and then learned to generate a new sample by adjusting the input SAR data to align with the alterations in the optical data. 


\subsubsection{Image Composition}

The task of image composition is primarily necessary for the detection of SAR targets in complex environments. The objective of this study is to generate a composition image that seamlessly blends a variety of SAR target patches, SAR scene images with complex background, and other source images, if available, in a logically consistent manner. The computer vision domain has extensively investigated and applied the concept in diverse applications, including digital photography and graphic design \cite{niu2021making}. The main steps include object placement, image blending, image harmonization and shadow generation. During the era of generative AI, the task of generative image composition has evolved into an integrated process that combines the aforementioned sub-tasks into a single unified model. There are a few SAR-related studies working on this direction \cite{sunDSDetLightweightDensely2021,kuangSemanticLayoutGuidedImageSynthesis2023,luanBCNetBackgroundConversion2024} which we introduced in the following. 

Sun et al. \cite{sunDSDetLightweightDensely2021} proposed a style embedded ship augmentation network (SEA) to seamlessly blend the ship target slices with the background images. Given the location masks, the generator could output the harmonized images. Kuang et al. \cite{kuangSemanticLayoutGuidedImageSynthesis2023} suggested a paradigm for collaborative SAR sample augmentation, aiming to achieve flexible and diversified high-quality sample augmentation. Initially, semantic layout was utilized to guide the generation of detection samples. Additionally, a Pix2Pix network was employed to effectively enhance the variety of backgrounds under the guidance of layout maps. Finally, progressive training strategy of the conditional diffusion model, together with a sample cleaning strategy was adopted for sample quality improvement. Luan et al. \cite{luanBCNetBackgroundConversion2024} introduced a SAR target background conversion network (BCNet) that integrated slices of targets of interest (TOI) with slices of large-scene. The TOI slices' background was transformed into large-scene background, while still maintaining the target's scattering characteristics. In this way, the generated data with background diversity and variability can be supplemented to the original dataset. 

\subsection{Towards Improving Data Quality}

Generative models are extensively employed in the domain of computer vision for various image reconstruction tasks, including denoising, deblurring, dehazing, and super-resolution, with the objective of enhancing the overall image quality. There are also some similar tasks in SAR community, such as SAR image despeckling, auto-focusing, super-resolution, and interference suppression \cite{weiCARNetEffectiveMethod2022,liuAFnetPAFnetFast2022}. However, there are certain peculiarities in the image quality deterioration process when it comes to the particular properties of SAR. 

\subsubsection{Despeckling}

The major difference of denoising application between SAR and optical data lies in the formation of noise, and it differs due to the nature of the sensors and the imaging processes involved. For SAR image, the thermal noise from the electronics, noise from the radar pulses or antenna, and the well-known speckle noise, could contaminate SAR signal and lead to degraded imaging result. Among them, the speckle noise has attracted most attention in recent years. It is a result of the random scattering of radar waves by the terrain or objects in the scene, and is influenced by radar system parameters, scene characteristics, and the imaging conditions. Speckle noise is spatially correlated and signal dependant as a result of SAR's coherent sensing nature; this can be expressed as a multiplicative model. On the other hand, optical image noise is often addictive. Fracastoro et al. have conducted a thorough survey regarding SAR image despeckling based on deep learning methods \cite{fracastoroDeepLearningMethods2021} so that we will not include too much in this field. Some selected representative work after 2022 are reviewed in this section, including diffusion model-based approaches. 

By utilizing the capabilities of CycleGANs, the proposed CycleSAR \cite{lattariCycleSARSARImage2023} regarded the despeckling task as an unpaired image-to-image translation task, enabling training without ground truth images. The integration of a Conditional VAE (CVAE) allowed for generating diverse speckled images, thus resulting in enhanced despeckling performance. Combing the similarity-based block-matching and noise referenced deep learning network, Wang et al. \cite{wangSARImageDespeckling2022} presented a new self-supervised speckling method based on auto-encoder. The proposed method exhibited both excellent despeckling performance and impressive generalization capacity. The proposed SAR-CAM \cite{koSARImageDespeckling2022} enhanced the performance of an encoder-decoder CNN architecture for SAR image despeckling by incorporating different attention modules. Furthermore, the multi-scale information were obtained through a context block at the minimum scale. 

Recently, the diffusion model-based SAR image despeckling have emerged. In SAR-DDPM \cite{pereraSARDespecklingUsing2023}, the despeckled images were acquired during the reverse process, where the added noise were predicted by a noise predictor conditioned on the speckled image. Furthermore, in order to improve the despeckling results, a novel inference approach utilizing cycle spinning was implemented. Another diffusion-based despeckling technique is R-DDPM \cite{hu2024sar}, which used overlapping region sampling in the reverse process to direct inverse diffusion, enabling excellent despeckling of SAR images without artifacts.

\begin{figure*}[!htbp]
    \centering
    \includegraphics[width=0.8\textwidth]{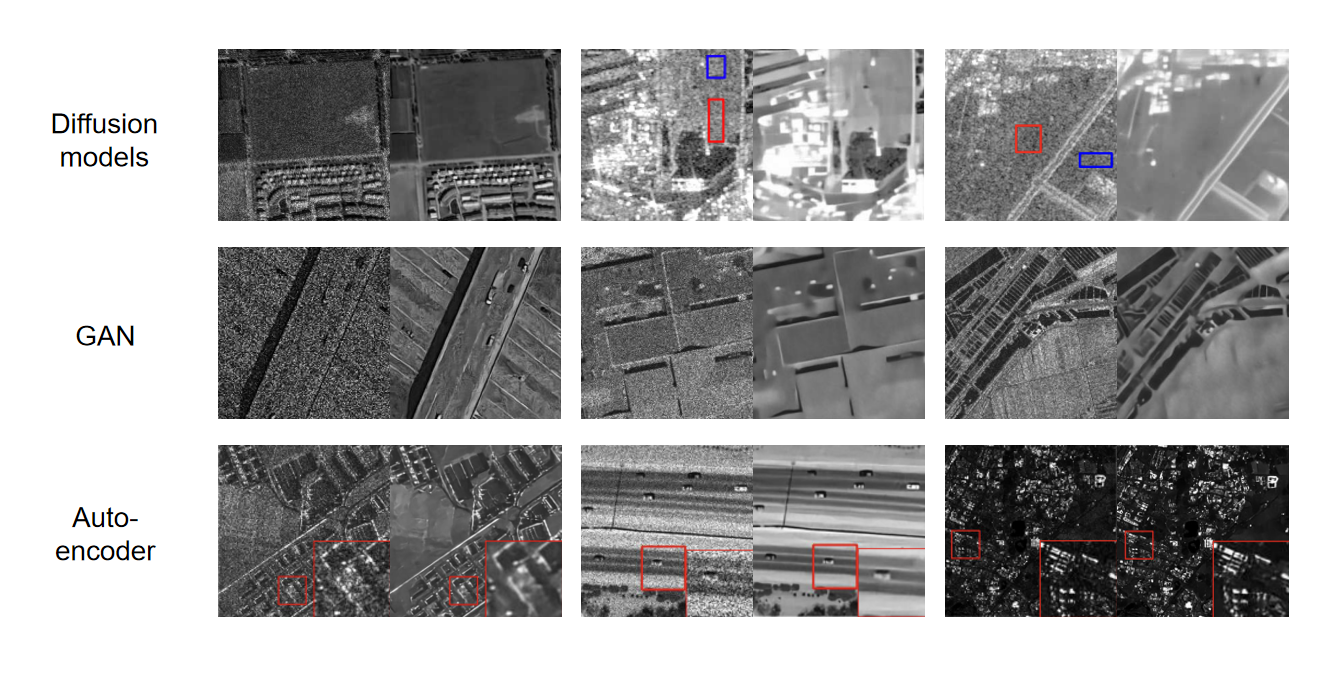}
    \caption{Comparison of different types of generative methods used for SAR despeckling in urban areas. The samples of diffusion models are from \cite{pereraSARDespecklingUsing2023}; the samples of GAN are from \cite{liu2020sar};  the samples of auto-encoders are from \cite{koSARImageDespeckling2022}.}
    \label{fig:compare}
\end{figure*}

In Fig. \ref{fig:compare}, we present despeckling results for urban areas using different types of models, including a diffusion model \cite{pereraSARDespecklingUsing2023}, GANs \cite{liu2020sar}, and auto-encoders \cite{koSARImageDespeckling2022}. These results demonstrate that all the generative models are effective in removing speckle noise from the images. The diffusion-based method demonstrated superior despeckling performance by more effectively preserving the fine structural details of the original images.

\subsubsection{Colorization}

In computer vision domain, the colorization aims to synthesize the color information for a gray-scale image, for applications such as historical photography recovery, archival restoration, and artistic manipulation. The single polarized SAR image is visually represented in grayscale, which can be less easily interpreted. Nevertheless, the task of colorization in the SAR domain may exhibit fundamental differences due to the unique imaging mechanism employed. The term "pseudo-color" SAR images is used to describe polarimetric SAR data with multiple channels, where polarimetric decomposition processing is performed to assign distinct polarimetric features to color channels. As a result, colorization would leads to different task definition for SAR.

\textbf{SAR-to-Optical Translation:} It is one of the most widely accepted definitions of SAR image colorization which pertains to the modality transfer problem. Compared with O2S task, SAR-to-Optical (S2O) is well-established and widely explored. The input SAR image is transferred to an optical image with true colors that represent the visual characteristics of objects. The contours, shapes, textures in optical features, are depicted in the generated optical image. This task will facilitate applications such as cloud removal, vegetation index reconstruction, and the construction of high temporal and spectral resolution time-series data, etc.

The researches focusing on this task include \cite{baiConditionalDiffusionSAR2024,shiBraininspiredApproachSARtooptical2024,zhangComparativeAnalysisEdge2021,yangSARtoopticalImageTranslation2022,niuImageTranslationHighresolution2021,leeCFCASETCoarsetoFineContextAware2023,zhangSARtoOpticalImageTranslation2024,weiCFRWDGANSARtoOpticalImage2023,wangSARtoOpticalImageTranslation2022,liDeepTranslationGAN2021,fuReciprocalTranslationSAR2021,liMultiscaleGenerativeAdversarial2022,guoEdgePreservingConvolutionalGenerative2021,yangFGGANFineGrainedGenerative2022,wangGeneratingHighQuality2018} and we will elaborate them in the following sections.

\textbf{SAR-to-Optical-Color Translation:} In this paper, we refer to another form of colorization, similar to S2O, as SAR-to-Optical-color (S2Oc) translation.  S2O translates the gray-scale SAR images to optical ones completely with optical features of objects including colors. On the other hand, the S2Oc translation process exclusively converts gray-scale pixels into colored ones, while preserving the primary structure of objects. Thus, by altering only the color information, the scattering characteristics of SAR remain unchanged. 

A few representative studies explored this task, such as \cite{wangGeneratingHighQuality2018,schmitt2018colorizing,shenBenchmarkingProtocolSAR2024}. Wang et al. \cite{wangGeneratingHighQuality2018} introduced a DCGAN-based method designed for despeckling and colorization progressively. In \cite{schmitt2018colorizing}, the creation of artificial color SAR images was initially undertaken as a result of the lack of available color SAR images for training purposes. This was achieved by fusing SAR and multi-spectral images. Furthermore, in order to produce multi-modal colorization hypotheses, both a variational autoencoder (VAE) and a mixture density network (MDN) were employed. Shen et al. \cite{shenBenchmarkingProtocolSAR2024} proposed a benchmarking protocol for SAR colorization. The authors introduced several baselines for SAR colorization and then a Pix2Pix-based network called cGAN4ColSAR was specially designed for the purpose of SAR colorization.

\textbf{Polarimetric SAR Image Generation: }In contrast to the aforementioned definitions, this term pertains to the process of inferring the polarimetric physical scattering characteristics from a gray-scale single polarimetric or dual-polarization SAR data. The extraction of physical information from polarized SAR images is more informative compared to single and dual polarization SAR data. From a theoretical standpoint, the problem can be characterized as an inverse and ill-posed one, as it is not possible to directly obtain signals in other polarization channels from a single polarization channel. By utilizing an AI model, it becomes feasible to address this issue to a certain degree. Once the objects are interpreted from single polarized SAR image and the characteristics such as geometry and surface roughness are identified, AI model can be employed to transfer the knowledge learned from physical model to generate the polarimetric physical scattering parameters of the objects. Several studies have identified the feasibility primarily.

Song et al. \cite{songRadarImageColorization2018} proposed the concept of "radar image colorization" of reconstructing the full-pol image from a non-full-pol SAR image to enable the current PolSAR image analysis methods can be applied to it. The "color" refers to polarimetric decomposition features, and the designed deep neural network aims to predict the polarimetric covariance matrix of each pixel. Based on this work, Deng et al. \cite{dengUrbanDamageLevelEstimation2024a,dengQuadPolSARData2023a} proposed to generate the pseudo-quad-pol SAR image from dual-pol SAR data which reconstruct the fully polarimetric information partially. The generated pseudo-quad-pol SAR images are then used for urban damage-level estimation based on polarimetric coherence pattern interpretation. 

Another related field is full-pol coherence matrix or covariance matrix reconstruction from compact polarimetric (CP) SAR with deep neural networks \cite{zhang2022compact}. Zhang et al. \cite{zhangPseudoQuadPolSimulation2022a} proposed a complex-valued dual-branch convolutional neural network (CV-DBCNN) to achieve the reconstruction of the pseudo quad-pol data from CP SAR image, and the method was evaluated with polarimetric decomposition to demonstrate its superiority in terms of the pseudo quad-pol data reconstruction and scattering mechanism preservation. Aghababaei et al. \cite{aghababaeiDeepLearningBasedPolarimetricData2023} proposed a deep learning based framework to reconstruct the full-pol information from typical dual-pol data. Specifically, the designed loss function accounted for different scattering properties of the polarimetric data, such as the covariance matrix, SPAN which refers to the image generated by the sum of the diagonal elements of the covariance matrix, and the statistical distribution of the eigenvalues of covariance matrix.

\subsection{Others}

In addition to the aforementioned applications, there exist several other less common applications related to generative models. We will provide a brief introduction to these applications below.

\textbf{Interference Suppression: }Radio frequency interference (RFI) is a critical issue in SAR imagery, as it can significantly degrade the quality and usability of the data. RFI refers to unwanted signals that contaminate the SAR signal and introduce noise into the image. Aiming at interference suppression using deep learning, Oyedare et al. \cite{oyedareInterferenceSuppressionUsing2022} proposed to review the literature in depth, in which the auto-encoder based models are considered as one of the popular approaches. In SAR image domain, Wei et al. \cite{WEI2022103019} proposed an auto-encoder based CARNet to address the SAR image interference suppression problem. The encoder–decoder architecture was designed to suppress interference and produce hierarchical-level feature maps for target information exchange.

\textbf{SAR Image Super-resolution: } It is a not well-defined task instead of that in optical image domain. Several studies adhere to the concept of enhancing spatial details, which aligns with the definition established for optical images \cite{shenResidualConvolutionalNeural2020,guSARImageSuperResolution2019,kongDMSCGANCGANBasedFramework2023}. The aim is to improve the visual quality of the low-resolution SAR images. Another definition posits that SAR image super-resolution involves the generation of a SAR image that closely resembles the output of a high-resolution sensor, based on the SAR image acquired from a different sensor with lower resolution \cite{ao2018dialectical}. The high-resolution SAR sensor, due to its wide bandwidth for transmitting waves and large synthesized aperture, is capable of producing SAR images with significantly distinct characteristics compared to low-resolution ones. According to this definition, SAR image super-resolution cannot be equated directly with the enhancement of visual quality.

Some selected work are briefly introduced. A residual convolutional neural network was proposed for polarimetric SAR image super-resolution in \cite{shenResidualConvolutionalNeural2020}. The interpolation was replaced by deconvolution to reduce the loss of precision and computational burden. The addition of PReLU helps to retain the negative information and enhance the accuracy. A complicated structure block was also specifically designed to preserve the complex features of PolSAR data. Some other researches achieved SAR image resolution based on generative adversarial network. For instance, Gu et al. \cite{guSARImageSuperResolution2019} proposed a noise-free GAN to generate high-resolution SAR images from low-resolution images. The despeckling network and the reconstruction network jointly constituted the generator and discriminator was used to distinguish real and fake high-resolution images. DMSC-GAN \cite{kongDMSCGANCGANBasedFramework2023} was another GAN-based SAR image super-resolution technique. In order to extract informative features, convolutional operations were combined with Deformable Multi-Head Self-Attention (DMSA) and a multi-scale feature extraction pyramid layer was also employed. Moreover, the perceptual loss and feature matching loss enabled the generator to obtain more comprehensive feedback from the discriminator.

\begin{figure*}[b]
    \centering
    \includegraphics[width=0.65\textwidth]{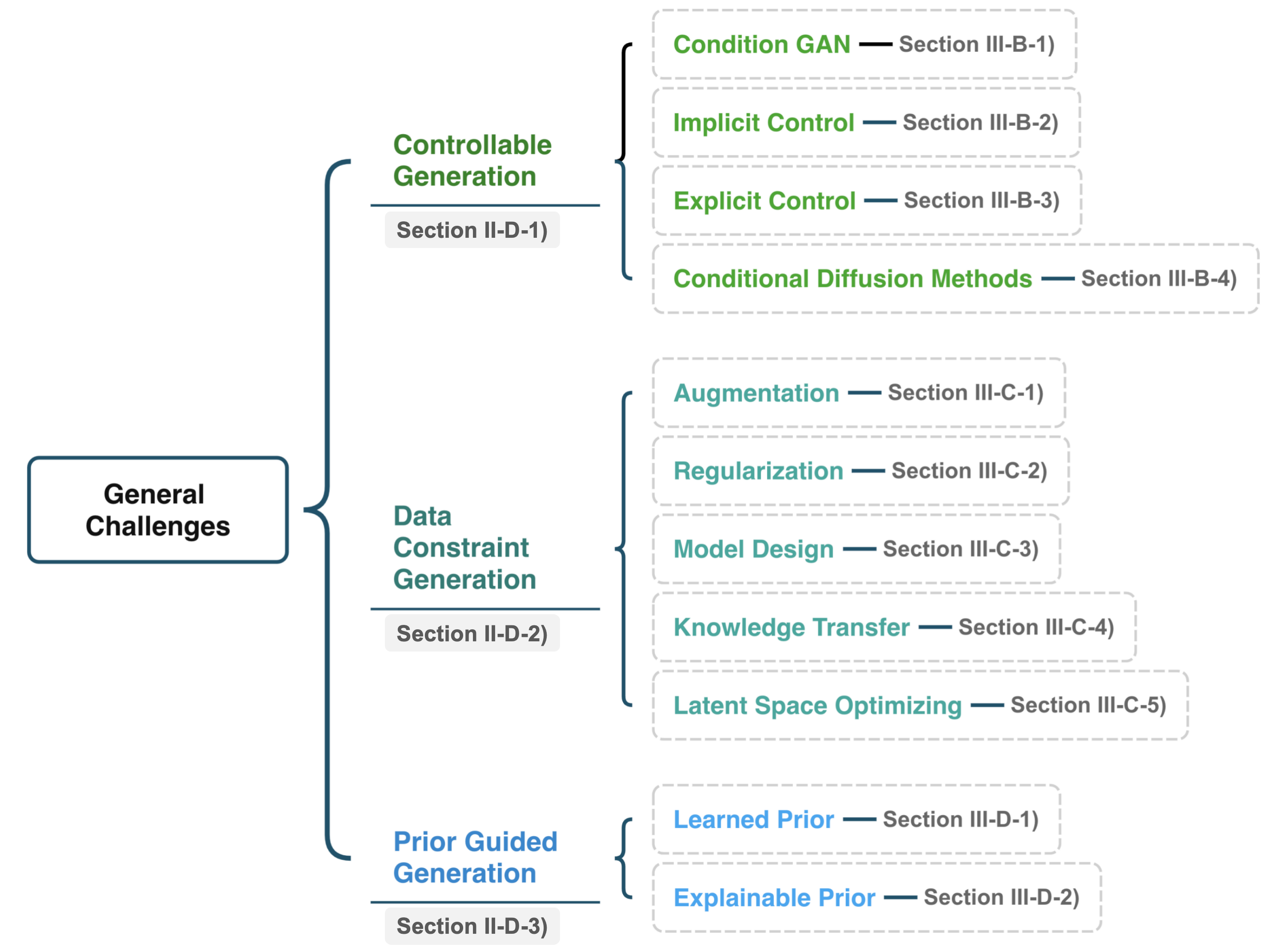}
    \caption{The general challenges in vision and SAR domain are summarized. The solutions and applications on SAR images can be referred in Section III.}
    \label{fig:challenge}
\end{figure*}

\textbf{Layout-to-Image Generation: } It pertains to the process of generating a variety of images based on a given layout, which includes the target information about the class and location. SARGAN \cite{juSARGANNovelSAR2023} was a GAN-based method proposed to generate various images for SAR ship detection task. The latent vector for each class was initially obtained through the target encoder and the generator can successfully produce diverse SAR ship images under the guidance of the constructed scene. Ship-Go \cite{zhang2024ship} is a specially designed instance-to-image multi-condition diffusion model to generate SAR images for object detection. Taking visual instances and environment prompt as conditions, Ship-Go can smoothly place the ship objects into the generated background of various environment types.


\subsection{General Challenges}

We summarize the general challenges in vision and SAR domain, including controllable generation, data constraint generation, and prior guided generation, as illustrated in Fig. \ref{fig:challenge}. The corresponding solutions will be elaborately introduced in the following section.

\subsubsection{Controllable Generation}

Controllable generation refers to the ability to generate images with specific attributes or features defined by users. It often requires the model to disentangle the underlying factors of variation in the data. If the model cannot separate these factors in the latent space, controlling individual attributes becomes difficult because changes in one attribute may inadvertently affect others. In SAR image domain, the most widely explored attributes are target category and azimuth angle \cite{giry2022sar,shiISAGANHighFidelityFullAzimuth2022,sunAttributeGuidedGenerativeAdversarial2023,ohPeaceGANGANBasedMultiTask2021,huFeatureLearningSAR2021,songSARImageRepresentation2019,songLearningGenerateSAR2022,guoCausalAdversarialAutoencoder2023,lei2024sar}. Besides, the background clutter distribution and the complex environment can be also a controlled factor for SAR image generation \cite{zhang2024ship}.



\subsubsection{Data Constraint Generation}

\begin{figure*}[!htbp]
    \centering
    \includegraphics[width=0.9\textwidth]{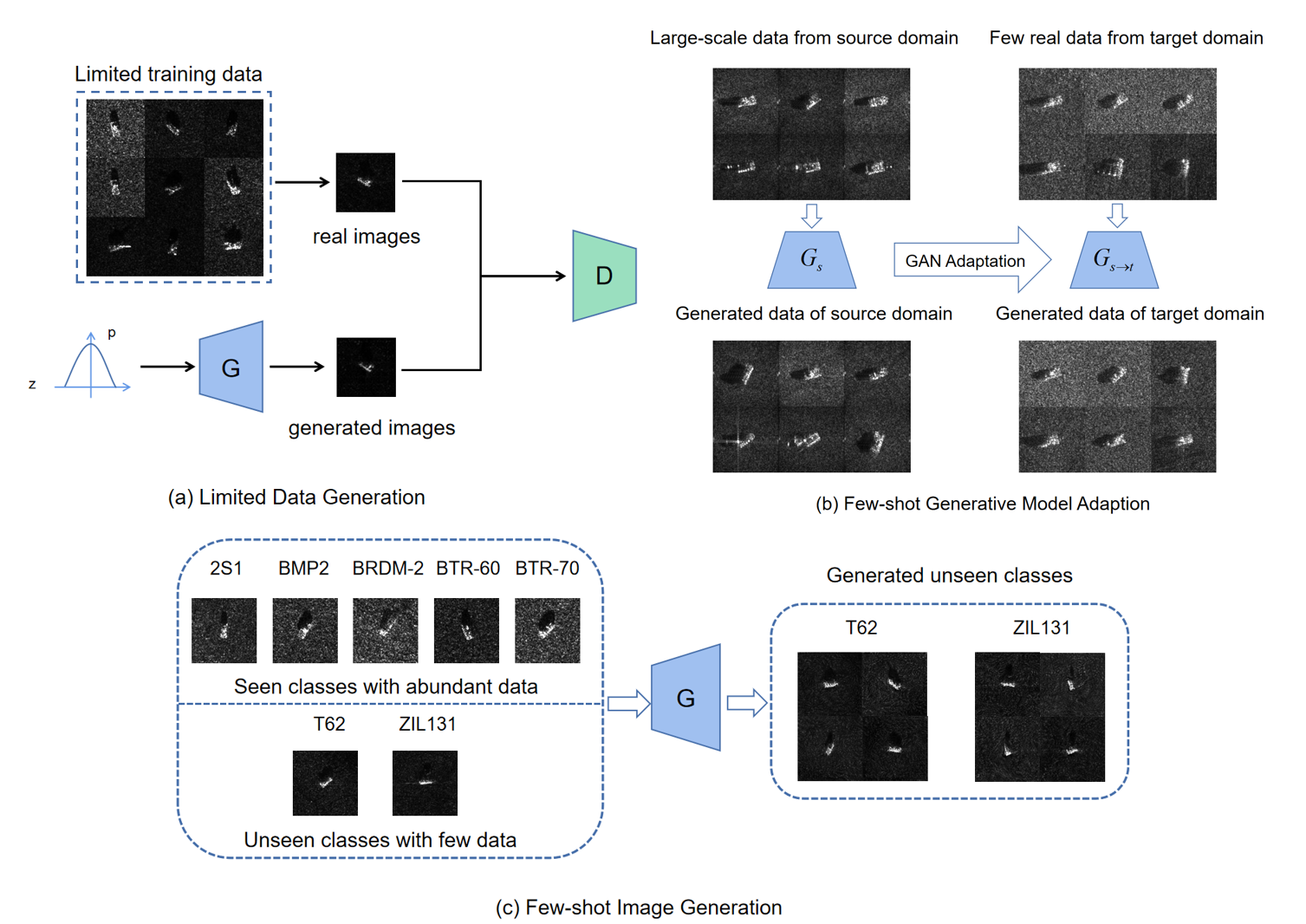}
    \caption{Data constraint generation settings for SAR.}
    \label{fig:few3t}
\end{figure*}

Data constraint is an important topic when data acquisition is challenging. Training a generative AI model inevitably requires sufficient data to achieve good performance, but lack of training samples is a main obstacle in real-world applications. Abdollahzadeh et al. \cite{abdollahzadeh2023survey} proposed a comprehensive literature review on generative modeling with limited data, few shots, and zero shot. Similarly, Yang et al. \cite{yang2023image} summarized image synthesis under limited data. In this paper, we conclude the data constraint challenge in generative modeling as the following three aspects based on representative computer vision tasks and the corresponding crucial SAR applications, as given in Fig. \ref{fig:few3t}.

\textbf{Limited Data: }It refers to traditional limited training data challenge, where the generative model is trained on insufficient data that would lead to mode collapse, overfitting, and unstable training. In SAR community, it is often occurs with limited training data due to the high cost of data acquirement compared with computer vision field. On the other hand, some specific applications, such as learning to generate SAR targets of different azimuth angles from sparse observations, also belong to this issue, as illustrated in Fig. \ref{fig:few3t} (a).

\textbf{Few-shot Generative Model Adaption: }It refers to transferring the knowledge of pre-trained generative AI models from large-scale source domains to target domains with limited data. The adapted generative model should not only inherit the generation ability of source model with invariant attributes, but also be aware of the target distribution to generate novel images. The similar application in SAR domain is presented in Fig. \ref{fig:few3t} (b). The generative model $\mathcal{M}_1$ is trained on SAR images obtained from sensor A, and it is capable to generate multi-view SAR targets controlled by azimuth angle. Given a few SAR images obtained from sensor B with different system parameters, the few-shot generative model adaption aim to adapt $\mathcal{M}_1$ to $\mathcal{M}_2$. Different sensor parameters would lead to domain drift such as various noise distribution. As a result, $\mathcal{M}_2$ should be adapted to generate SAR images cross the sensors.

\textbf{Few-shot Image Generation: }It aims to generate the images of unseen classes with few-shot samples based on training with sufficient samples of seen classes. The definition follows few-shot learning with N-way-K-shot. We summarize a typical application scenario in SAR domain, as shown in Fig. \ref{fig:few3t} (c). Assume there are $M$ classes with sufficient SAR images obtained from complete observation angles. In addition, another $N$ classes only contain $K$ samples each with sparse observation. Given these training data, the few-shot image generation task aims to enable the model to generate diverse samples in unseen classes. 

\subsubsection{Prior Guided Generation}

For image reconstruction task, such as inpainting, denoising, and super-resolution, priors indicate some empirical knowledge or objective laws that characterize the image properties and can be helpful to constrain the optimization space. How to extract or represent the prior from data or knowledge, and how to effectively use them to guide the generation, have become a general challenge for both computer vision and SAR applications. The image degradation model, for example, is one of the well-known priors for image restoration. Although different degradation models are employed by SAR and optical images, the solutions can be learned from each other.

\section{Generative AI Methods}

This section aims to introduce the popular deep generative models in recent years and review the related SAR applications based on them. First, the basic generative AI methods are investigated and the summary of literature is given in Table \ref{tab:basicmodel}. Then, a broad range of approaches in SAR and computer vision domain addressing the general challenges are reviewed and the summary can be found in Table \ref{tab:challenges}.

\begin{table*}[h!]
    \centering
    \caption{The representative studies in SAR and computer vision domain that were developed based on different basic models are summarized.}
    \label{tab:basicmodel}
    \begin{tabular}{ccccccc}
    \toprule
    \textbf{\makecell[c]{Basic \\ Model}} & \textbf{AE} &\textbf{DCGAN} &\multicolumn{2}{c}{\textbf{cGAN}} &\textbf{ACGAN} &\textbf{InfoGAN} \\
    \midrule
    \textbf{SAR} &\cite{huangAsymmetricTrainingGenerative2022,huFeatureLearningSAR2021,songLearningGenerateSAR2022,songSARImageRepresentation2019,wangSyntheticApertureRadar2019,guoCausalAdversarialAutoencoder2023}
    &\cite{wangSARTargetImage2022,gaoDeepConvolutionalGenerative2018}
    &\multicolumn{2}{c}{\makecell[c]{\cite{guoSARImageData2022,juSARImageGeneration2024,huangHighResolutionSAR2019,qinTargetSARImage2022,duHighQualityMulticategorySAR2022,zhuLIMEBasedDataSelection2022,daiCVGANCrossViewSAR2023,caoLDGANSyntheticAperture2020,caoDemandDrivenSARTarget2022,shiISAGANHighFidelityFullAzimuth2022}\\\cite{zhangComparativeAnalysisEdge2021,yangSARtoopticalImageTranslation2022,niuImageTranslationHighresolution2021,leeCFCASETCoarsetoFineContextAware2023,zhangSARtoOpticalImageTranslation2024,guoSyntheticApertureRadar2017,liSARtoOpticalImageTranslation2020,noaturnesAtrousCGANSAR2022,guoSar2colorLearningImaging2022,wangHybridCGANCoupling2022}}}
    &\cite{zouMWACGANGeneratingMultiscale2020,kongSARTargetRecognition2021,pengGenerationSARImages2022,sunAttributeGuidedGenerativeAdversarial2023,ohPeaceGANGANBasedMultiTask2021}
    &\cite{fengInterpretationLatentCodes2023a} \\
    \textbf{CV} &\cite{rumelhart1986parallel,kingma2013auto,makhzani2015adversarial}  
    &\cite{radford2015unsupervised}
    &\multicolumn{2}{c}{\makecell[c]{\cite{mirza2014conditional}}}
    &\cite{odena2017conditional}
    &\cite{chen2016infogan} \\
    \midrule
    \textbf{\makecell[c]{Basic \\ Model}} &\textbf{PGGAN} &\textbf{StyleGAN} &\textbf{Pix2Pix} &\textbf{CycleGAN} &\textbf{\makecell[c]{Diffusion \\ Model}} &\textbf{NeRF} \\
    \midrule
    \textbf{SAR}
    &\cite{luo2020synthetic,songSARImageRepresentation2019,songLearningGenerateSAR2022}
    &\cite{giry2022sar,guoCausalAdversarialAutoencoder2023}
    &\makecell[c]{\cite{sunDSDetLightweightDensely2021,juSARGANNovelSAR2023,kuangSemanticLayoutGuidedImageSynthesis2023} \\ \cite{weiCFRWDGANSARtoOpticalImage2023,wangSARtoOpticalImageTranslation2022}}
    &\cite{liDeepTranslationGAN2021,fuReciprocalTranslationSAR2021,liMultiscaleGenerativeAdversarial2022,guoEdgePreservingConvolutionalGenerative2021,yangFGGANFineGrainedGenerative2022,fuentesreyesSARtoOpticalImageTranslation2019}
    &\cite{shiBraininspiredApproachSARtooptical2024,baiConditionalDiffusionSAR2024,xu2023recognizer,zhang2024ship} &\cite{lei2024sar,zhangCircularSARIncoherent2023,ehret2024radar,liu2023ranerf,deng2024isar} \\
    \textbf{CV}
    &\cite{karras2017progressive}
    &\cite{karras2019style,karrasAnalyzingImprovingImage2020,karras2021alias}
    &\makecell[c]{\cite{isola2017image}}
    &\cite{zhuUnpairedImagetoImageTranslation2017}
    &\cite{ho2020denoising} 
    &\cite{mildenhallNeRFRepresentingScenes2022} \\
    \bottomrule
    \end{tabular}
\end{table*}

\subsection{Basic Models}

\subsubsection{Auto-Encoders}

\begin{figure*}[!htbp]
    \centering
    \includegraphics[width=1.0\textwidth]{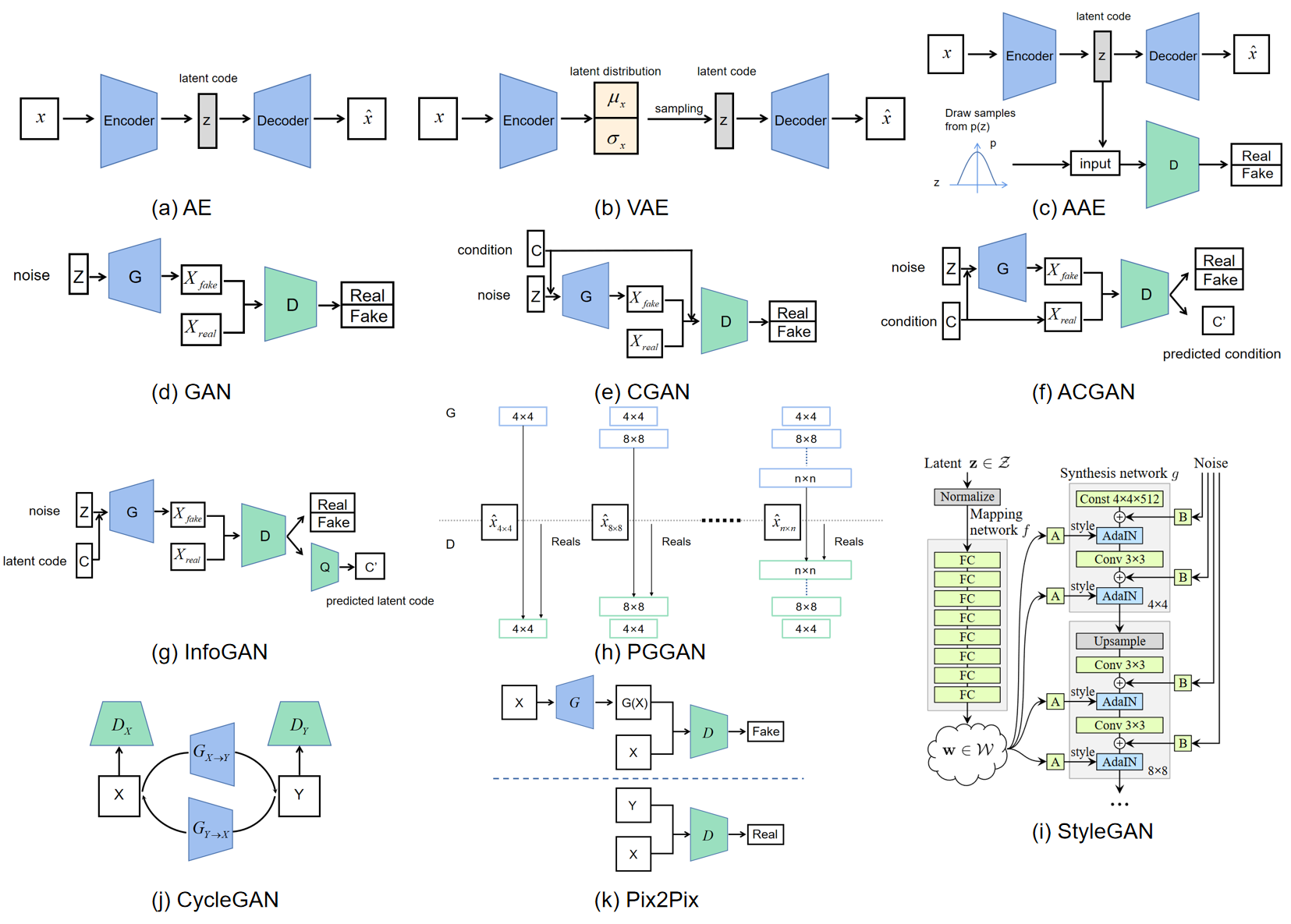}
    \caption{The basic models of Auto-Encoders and GANs}
    \label{fig:basicmodel}
\end{figure*}

Auto-encoders primarily concentrate on transforming input data into a space with less dimensions and subsequently reverting it back to the original space through decoding. The decoder component can be seen as a generator. The popular AE-based generative models contain variational auto-encoder (VAE) and adversarial auto-encoder (AAE). The brief illustration can be found in Fig. \ref{fig:basicmodel}.


VAE and AAE are briefly introduced in this section. The illustrations of the AE-based basic models are given in Fig. \ref{fig:basicmodel}. More architectures employ the combination of VAE and GAN, where the discriminator in GAN is introduced into VAE model to improve the quality of generated image, such as IntroVAE \cite{huang2018introvae}.

\textbf{Variational Auto-Encoder (VAE) \cite{kingma2013auto}: }The VAE is a probabilistic model that consists of an encoder and a decoder, which maps input data to a latent space and back to the input space, respectively. The key innovation of the VAE is the introduction of a prior distribution on the latent space and the use of variational inference to approximate the posterior distribution of the latent space given the input data. This approach allows the VAE to learn a distribution of latent vectors, which can be sampled to generate new data. The latent space representation is important for data generation and the disentanglement learning of the latent code allows interpretable and controllable generation \cite{yangCausalVAEDisentangledRepresentation2021}.

\textbf{Adversarial Auto-Encoder (AAE) \cite{makhzani2015adversarial}:} AAE is another form of probabilistic autoencoder that incorporates the adversarial training framework from Generative Adversarial Networks (GANs) to conduct variational inference by aligning the aggregated posterior of the hidden code vector of the autoencoder with a specified prior distribution, ensuring that the generation of samples from any portion of the prior space yields meaningful results. In Adversarial Autoencoders, in order to minimize the difference between the the prior and the approximated posterior, the KL-divergence in the VAE loss function is substituted with a adversarial training loss, enabling AAE to impose more complicated distributions.

\textbf{Applications on SAR: }
Some preliminary studies explored the potential of VAE model for SAR image generation. Wang et al. \cite{wangSyntheticApertureRadar2019} applied the Wasserstein autoencoder for the first time to generate SAR targets and improved the reconstruction loss by considering the speckle influence of SAR image. The combinations of auto-encoders and GAN have attracted attention in the following studies. More combined model, such as VAE-WGANGP \cite{huangAsymmetricTrainingGenerative2022} and CVAE-GAN \cite{huFeatureLearningSAR2021}, were introduced in this field to improve the quality of generated SAR images. Recently, a causal AAE model was proposed \cite{guoCausalAdversarialAutoencoder2023} which the auto-encoder was combined with StyleGAN model. The AAE architecture is used for random sampling which is then fed into the causal decoder with disentangled factors.



\subsubsection{Generative Adversarial Networks}

By implementing a competitive framework comprised of two neural networks—the discriminator and the generator—Generative Adversarial Networks (GANs) have significantly altered the domain of image generation. GANs have effectively been utilized in several areas, including image synthesis, style transfer, and photo editing. Most SAR image generation algorithms are developed using GANs and its several forms, including conditional GANs (cGAN), Wasserstein GANs (WGANs), Deep Convolutional GANs (DCGANs), InfoGANs, and others \cite{wangSyntheticApertureRadar2019}. The brief illustrations can be found in Fig. \ref{fig:basicmodel}.


\textbf{GAN \cite{goodfellow2014generativeadversarialnetworks}:} GANs are composed of two neural networks—a generator and a discriminator. The generator tries to create data (e.g., images) that come from the same distribution as the training data, while the discriminator tries to distinguish between real and generated data. The two networks are trained simultaneously in a competitive zero-sum game: the generator tries to get better at fooling the discriminator, and the discriminator tries to get better at detecting fakes.

\textbf{DCGAN \cite{radford2015unsupervised}:} Deep Convolutional Generative Adversarial Network (DCGAN) is a specific architecture of GANs that uses convolutional layers in both the generator and discriminator. It introduced a set of guidelines for designing GANs that are more stable and capable of learning realistic image distributions. DCGANs have been particularly successful in generating high-quality images.

\textbf{cGAN \cite{mirza2014conditional}:} Conditional Generative Adversarial Network (cGAN) is an extension of GANs where both the generator and discriminator are conditioned on some additional information (e.g., class labels or text descriptions). This conditioning allows the model to generate images with specific attributes or according to particular input conditions.

\textbf{InfoGAN \cite{chen2016infogan}:} Information Maximizing Generative Adversarial Nets (InfoGAN) is a specialized architecture of GANs designed for the purpose of exerting more control over the generation direction of GANs. The latent codes are introduced for further disentangling the input noise. A strong correlation between latent codes and the attributes can be established in an unsupervised manner by maximizing their mutual information during training process.

\textbf{ACGAN \cite{odena2017conditional}:} Auxiliary Classifier GAN (ACGAN) is a variant of cGAN that introduces an auxiliary classifier along with the generator and discriminator. The classifier helps guide the generator to produce images that not only look real but also correspond to the correct class label. This additional guidance can improve the quality and diversity of the generated images.


\textbf{PGGAN \cite{karras2017progressive}:} Progressive Growing of GANs (PGGAN) is an approach to train GANs that starts with generating very small images and gradually increases the resolution. At each resolution, the model adds new layers to the existing network without starting the training process from scratch. This progressive training helps stabilize the learning process and allows GANs to generate high-resolution images.

\textbf{StyleGAN \cite{karras2019style}:} StyleGAN is a GAN architecture that allows for more control over the generation process by separating the generation of high-level attributes (like pose and identity) from low-level details (like texture). It achieves this by introducing intermediate latent spaces that control different “styles” of the generated images. StyleGAN can produce images with incredible detail and realism. StyleGAN2 \cite{karrasAnalyzingImprovingImage2020} was then introduced by Karras et al. as an improvement over the former StyleGAN version. It addresses the limitations of identifying distinct artifacts in the generated images by modifying the generator normalization, replacing the PGGAN architecture with the MSG-GAN architecture \cite{karnewarMSGGANMultiScaleGradients2020}, and implementing generator regularization to promote optimal conditioning in the mapping from latent codes to images. In order to solve the problem of texture sticking further improve the quality of generated images, Karras et al. propsed StyleGAN3 \cite{karras2021alias}, offering a systematic resolution to suppressing aliasing, enabling the model to incorporate a more natural hierarchical refining process. 

\textbf{Pix2Pix \cite{isola2017image}:} Pix2Pix is a conditional GAN used for image-to-image translation tasks. It takes a pair of images as input (e.g., an input image and a corresponding output image) and learns a mapping from the input image to the output image. This model is useful for applications like semantic segmentation, photo enhancement, and style transfer.

\textbf{CycleGAN \cite{zhuUnpairedImagetoImageTranslation2017}:} CycleGAN is an approach for learning to translate an image from a source domain to a target domain in the absence of paired examples. In order to make the mapping from source domain to target domain becomes highly constrained, cycle consistency loss was introduced, ensuring that the original image can return to its original state after undergoing two conversions.



\textbf{Applications on Remote Sensing: }
GANs have found extensive application in remote sensing, demonstrating notable success in tasks such as augmenting data, change detection, spatial-temporal-spectral fusion, image super-resolution, etc. The existing review literature have demonstrate the current research progress in the field of remote sensing \cite{jozdani2022review,wang2023review}. We only list some up-to-date researches for illustration. For instance, to address the challenge of recovering missing details in large-factor image super-resolution, spectra-guided generative adversarial networks (SpecGANs) \cite{meng2022large} were proposed, utilizing additional hyperspectral images (HSIs) as conditional inputs to the GAN model. Liu et al. \cite{liu2022physics} further explored this area by using a GAN framework that considers both imaging mechanisms and spectral mixing to translate RGB images into high-resolution hyperspectral images and subpixel ground-truth annotations. Conducted Semantic Embedding GAN (CSEBGAN) \cite{wang2023remote} was introduced to generate semantic-controllable remote sensing images, which can be used to augment training datasets and enhance the performance of downstream remote sensing image segmentation tasks.

\textbf{Applications on SAR: }
According to the basic models studies applied, a majority of work concerning SAR target generation are based on cGAN \cite{mirza2014conditional}. Among them, the class label is commonly used as the input condition \cite{guoSARImageData2022,juSARImageGeneration2024,huangHighResolutionSAR2019,qinTargetSARImage2022,duHighQualityMulticategorySAR2022,zhuLIMEBasedDataSelection2022,caoLDGANSyntheticAperture2020,caoDemandDrivenSARTarget2022}. The azimuth angle, as an important factor, is also considered. Shi et al. \cite{shiISAGANHighFidelityFullAzimuth2022} applied the cGAN with projection discrimination \cite{miyato2018cgans} to encode the angle information as condition. Besides, some other researches used an image as the conditional input to conduct the SAR-to-Optical translation \cite{zhangComparativeAnalysisEdge2021,yangSARtoopticalImageTranslation2022,niuImageTranslationHighresolution2021,leeCFCASETCoarsetoFineContextAware2023,zhangSARtoOpticalImageTranslation2024}, as well as angle interpolation of SAR target \cite{daiCVGANCrossViewSAR2023}. Literature \cite{zouMWACGANGeneratingMultiscale2020,kongSARTargetRecognition2021,pengGenerationSARImages2022,sunAttributeGuidedGenerativeAdversarial2023,ohPeaceGANGANBasedMultiTask2021} developed SAR image generation methods based on ACGAN \cite{odena2017conditional}, where the auxiliary classifier was used to estimate the class label \cite{zouMWACGANGeneratingMultiscale2020,kongSARTargetRecognition2021} and angle \cite{pengGenerationSARImages2022,sunAttributeGuidedGenerativeAdversarial2023,ohPeaceGANGANBasedMultiTask2021} of generated SAR image. Feng et al. \cite{fengInterpretationLatentCodes2023a} proposed a method based on InfoGAN \cite{chen2016infogan}, where the input noise can be disentangled and interpreted by establishing a mutual information based optimization to correlate the latent codes and image properties. In this way, the relation between attributes and latent codes can be interpreted and the generated images can be interpretably manipulated by changing the codes. Due to the stabilizing training of PGGAN \cite{karras2017progressive}, it was also applied for generating high quality SAR images in literature \cite{luo2020synthetic,songSARImageRepresentation2019,songLearningGenerateSAR2022}. StyleGAN and its variants \cite{karras2019style,karrasAnalyzingImprovingImage2020,karras2021alias} were utilized in literature \cite{giry2022sar,guoCausalAdversarialAutoencoder2023} where the target attributes of class label and angles were separately represented with a mapping network to $\mathcal{W}$ space. The former introduced researches mainly focus on SAR target image generation based on class label and angle values. Pix2Pix \cite{isola2017image}, as a specific cGAN for image-to-image translation, is widely applied in SAR image composition \cite{sunDSDetLightweightDensely2021,kuangSemanticLayoutGuidedImageSynthesis2023,luanBCNetBackgroundConversion2024} and SAR-to-Optical translation \cite{weiCFRWDGANSARtoOpticalImage2023,wangSARtoOpticalImageTranslation2022}. Owing to the cycle consistency loss which enforces an image translated from one domain as close as possible to the one translated back to its original domain, CycleGAN \cite{zhuUnpairedImagetoImageTranslation2017} was generally applied for SAR-to-Optical or Optical-to-SAR translation \cite{liDeepTranslationGAN2021,fuReciprocalTranslationSAR2021,liMultiscaleGenerativeAdversarial2022,guoEdgePreservingConvolutionalGenerative2021,yangFGGANFineGrainedGenerative2022}.


\subsubsection{Diffusion Model}

Diffusion models are a class of probabilistic generative models that introduce noise to data step by step and subsequently learn to reverse this process for the generation of novel samples.

\textbf{Denoising Diffusion Probabilistic Models (DDPMs): }DDPMs \cite{ho2020denoising} are a class of latent variable models inspired by concepts from nonequilibrium thermodynamics, utilizing two coupled Markov chains: a forward Markov chain and a reverse Markov chain. In the forward diffusion process, the input data is progressively perturbed by the incremental addition of Gaussian noise across multiple steps. Conversely, during the reverse process, the model is tasked with reconstructing the original input data by learning to iteratively reverse the diffusion process, restoring the data step by step.

In more detail, the forward process involves progressively adding Gaussian noise to the data until it converges to pure random noise. The outcome at each time-step $t$ depends on the outcome at the previous time-step $t-1$. As $t$ increases, the image progressively degrades, approaching a state of pure noise. Mathematically, the forward process for the raw data adds Gaussian noise at each step according to the following equations:

\begin{equation}
    q(x_{1:T}|x_0)=\prod_{t=1}^{T}q(x_t|x_{t-1})
\end{equation}
\begin{equation}
    q(x_t|x_{t-1})=N(x_t;\sqrt{1-\beta_t}x_{t-1},\beta_tI)
\end{equation}
where $x_t$ represents the data at time step $t$, $x_{1:T}$ denotes the sequence of data from time step $1$ to $T$, and $\beta_t$ is the noise coefficient at time step $t$, which controls the magnitude of the added noise.

The reverse process is a denoising operation that inverses the forward process, restoring the image from Gaussian noise by approximating the posterior distribution. This reverse process can be modeled as a Markov chain, expressed as:

\begin{equation}
p_{\theta}(x_{0:T})=p(x_T)\prod_{t=1}^{T} p_{\theta}(x_{t-1}|x_t)
\end{equation}

\begin{equation}
p_{\theta}(x_{t-1} \mid x_t) = \mathcal{N}\left(x_{t-1}; \mu_{\theta}(x_t, t), \Sigma_{\theta}\left(x_t, t\right)\right)
\end{equation}

where \( \mu_{\theta} \) represents the mean, and \( \Sigma_{\theta} \) denotes the covariance, both of which are parameterized by \( \theta \). The objective of DDPM is to minimize the difference between the true noise \( \epsilon \) and the noise predicted by the model \( \epsilon_{\theta} \), which can be formulated as:

\begin{equation}
   L = \mathrm{E}_{x, \epsilon\sim\mathcal{N}(0,1), t}\left[\|\epsilon - \epsilon_{\theta}(x_t, t)\|_2^2\right]
\end{equation}
It measures the mean squared error between the actual noise and the model's predicted noise, guiding the model to accurately reverse the diffusion process.

\begin{figure*}[!htbp]
    \centering
    \includegraphics[width=0.8\textwidth]{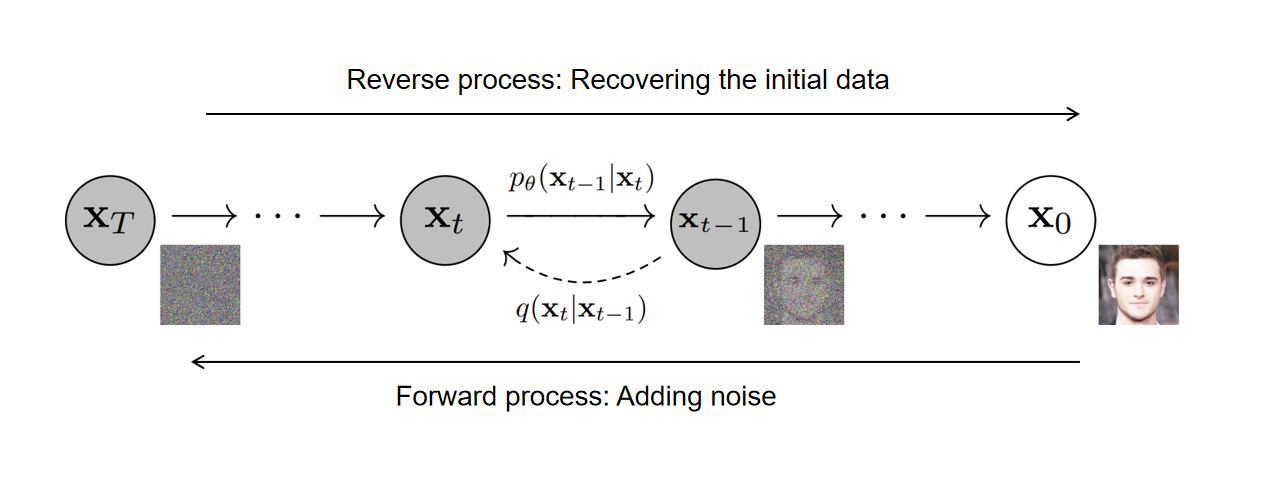}
    \caption{The principle of DDPM \cite{ho2020denoising}.}
    \label{fig:diffusion}
\end{figure*}

Diffusion models have recently been employed to address a variety of complex real-world challenges due to their adaptability and robust generative capabilities. One of the most challenging tasks is text-to-image generation, where the goal is to produce an image based on a given descriptive text. DALL$\cdot$E 2 proposes a two-step approach: first, a model generates a CLIP-based image representation from a text description; then, a diffusion-based decoder generates an image from this representation. Imagen, on the other hand, utilizes a text sequence encoder and a series of diffusion models to produce high-resolution images, relying heavily on the text embeddings generated by the encoder. Furthermore, pre-trained text-to-image diffusion models can be leveraged for more intricate or detailed manipulation of generated images. For example, ControlNet is designed to control large pre-trained diffusion models by incorporating additional semantic information, such as edge maps, segmentation maps, keypoints, surface normals, and depth.

\textbf{Applications on Remote Sensing: }
Recently, diffusion models have garnered attention for their applications in remote sensing data. For instance, DiffusionSat \cite{khanna2023diffusionsat} is an advanced model designed for satellite image generation, capable of producing high-resolution images from numerical metadata and text. It incorporates a specially designed 3D-conditioning extension, enabling it to achieve state-of-the-art performance in super-resolution, temporal generation, and inpainting. Another notable diffusion-based model, MetaEarth \cite{yu2024metaearth}, employs a self-cascading generative framework that uses resolution guidance to generate images at varying geographical resolutions for a given region. This process begins with low-resolution images and progressively refines them to high-resolution, with each stage influenced by the low-resolution outputs from the previous stage and their spatial resolution. Furthermore, diffusion models have shown promise in addressing cloud removal tasks. For example, DiffCR \cite{zou2024diffcr} utilizes conditional guided diffusion combined with deep convolutional networks to effectively remove clouds from optical satellite data, yielding high-performance results. Additionally, Conditional DDPM models have demonstrated the capability to generate high-quality, diverse satellite images that accurately correspond to input semantic maps \cite{baghirli2023satdm}.

\begin{figure*}[!htbp]
    \centering
    \includegraphics[width=0.9\textwidth]{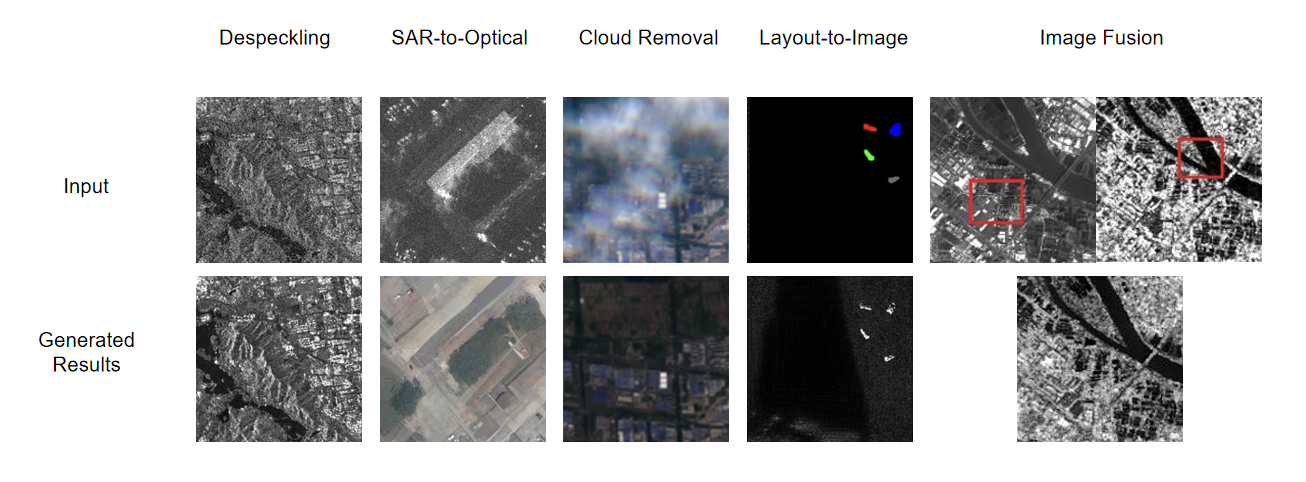}
    \caption{Illustrations of different applications of diffusion models in SAR. The samples are from \cite{hu2024sar,baiConditionalDiffusionSAR2024,jing2023denoising,kuangSemanticLayoutGuidedImageSynthesis2023,zhang2024variational} respectively. }
    \label{fig:dr}
\end{figure*}

\textbf{Applications on SAR: }Diffusion models have shown initial potentials on various applications of SAR, such as despeckling \cite{hu2024sar,pereraSARDespecklingUsing2023,guha2023sddpm, xiao2023unsupervised}, SAR-to-Optical image translation, cloud removal assisted with SAR images \cite{jing2023denoising}, data augmentation \cite{xu2023recognizer,kuangSemanticLayoutGuidedImageSynthesis2023}, and multi-source image fusion. R-DDPM \cite{hu2024sar} enables the flexible removal of speckle noise from SAR images across various scales within a single training session. It effectively mitigates artefacts in fused SAR images through region-guided inverse sampling. Another diffusion-based approach, SAR-DDPM \cite{pereraSARDespecklingUsing2023}, addresses speckle noise by leveraging estimates derived from multiple cycle-spinning techniques. Based on diffusion models, the following two studies \cite{baiConditionalDiffusionSAR2024,shiBraininspiredApproachSARtooptical2024} achieved SAR-to-Optical translation by using SAR images as conditional constraints during the sampling process. DDPM-CR \cite{jing2023denoising} was developed to remove clouds by using both cloudy optical and supplementary SAR images as inputs of diffusion model. This method effectively extracts features that contain valuable information for recovering lost contents. In the realm of data augmentation, Xu et al. \cite{xu2023recognizer} leveraged DDPM to generate additional target images, thereby supplementing limited datasets and enhancing target recognition performance. Kuang et al. \cite{kuangSemanticLayoutGuidedImageSynthesis2023} proposed a collaborative SAR sample augmentation paradigm aimed at achieving flexible and diverse high-quality sample augmentation through synthetic generation. Additionally, diffusion models have been employed to fuse optical imagery with synthetic aperture radar data, integrating crucial information from multiple sources \cite{zhang2024variational}. Some examples are illustrated in Fig. \ref{fig:dr}.


\subsubsection{Neural Radiance Field}

Even Neural radiance fields (NeRF) is not a typical generative model like GANs and Diffusion models where the new data is generated from a learned distribution, often starting from random noise, it does share some characteristics with GenAI models. For example, its use of neural networks to model complex data and its ability to generate novel-view images. In many researches, NeRF has been combined with GANs and diffusion models to achieve better results. Lately, some remote sensing generation tasks also applied it for novel-view scene generation. That's why we still introduce NeRF as the basic model.

\begin{figure*}[!htbp]
    \centering
    \includegraphics[width=0.9\textwidth]{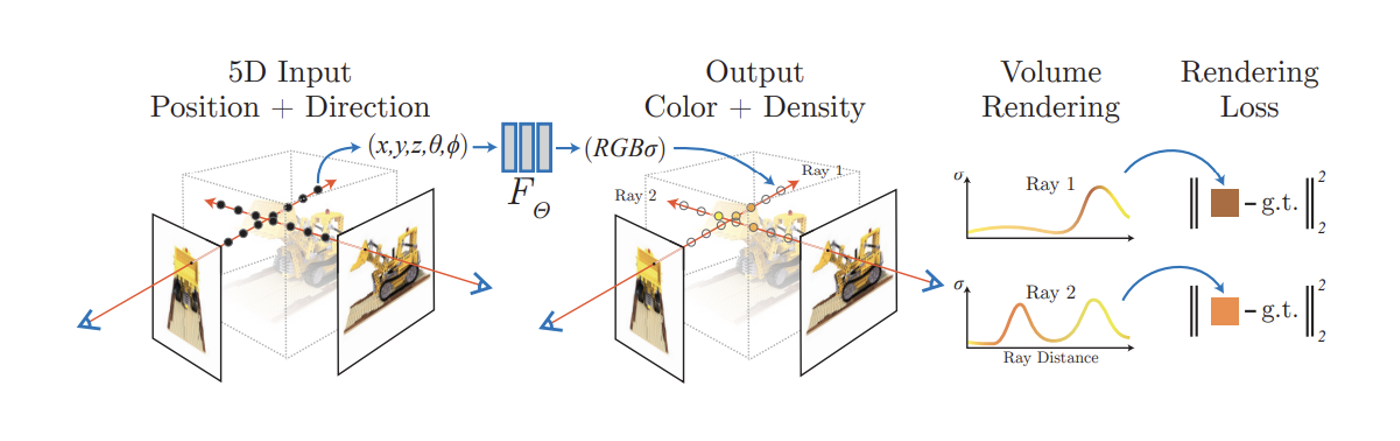}
    \caption{The overview of NeRF from \cite{mildenhallNeRFRepresentingScenes2022}. }
    \label{fig:nerf}
\end{figure*}

NeRF is a technique for representing 3D scenes and synthesizing novel views using a fully-connected deep neural network, as illustrated in Fig. \ref{fig:nerf}. The core idea of NeRF is to represent a scene as a continuous volumetric function that maps spatial coordinates \(\mathbf{x} = (x, y, z)\) and viewing direction \(\mathbf{d} = (\theta, \phi)\) to the RGB color \(\mathbf{c} = (r, g, b)\) and volume density \(\sigma\). This function is denoted as \(F_\theta\) and is parameterized by the network's weights \(\theta\):

\begin{equation}
  F_\theta : (\mathbf{x}, \mathbf{d}) \rightarrow (\mathbf{c}, \sigma)  
\end{equation}

To render an image, NeRF samples points along rays that pass through each pixel in the image plane. For a given ray, parameterized by \(\mathbf{r}(t) = \mathbf{o} + t\mathbf{d}\), where \(\mathbf{o}\) is the camera origin and \(t\) is the distance along the ray, NeRF predicts the color and density at various points along the ray. The final pixel color is obtained by accumulating the contributions of all points along the ray using volume rendering, which is expressed as:

\begin{equation}
\hat{\mathbf{C}}(\mathbf{r}) = \sum_{i=1}^{N} T_i \alpha_i \mathbf{c}_i
\end{equation}

where \(T_i = \exp\left(-\sum_{j=1}^{i-1} \sigma_j \Delta_j\right)\) is the accumulated transmittance, \(\alpha_i = 1 - \exp(-\sigma_i \Delta_i)\) is the opacity, and \(\Delta_j\) is the distance between consecutive sample points along the ray. The optimization of the network involves minimizing the difference between the rendered pixel colors \(\hat{\mathbf{C}}(\mathbf{r})\) and the ground truth colors \(\mathbf{C}(\mathbf{r})\) from the training images using a loss function like mean squared error:

\begin{equation}
\mathcal{L}(\theta) = \sum_{\mathbf{r} \in \mathcal{R}} \|\hat{\mathbf{C}}(\mathbf{r}) - \mathbf{C}(\mathbf{r})\|^2
\end{equation}

By optimizing this loss, NeRF learns a volumetric scene representation that can be used to synthesize photorealistic images from novel viewpoints \cite{mildenhallNeRFRepresentingScenes2022}.

\textbf{Applications on Remote Sensing: }
NeRF has been increasingly utilized by researchers for novel view synthesis in remote sensing. The proposed FReSNeRF \cite{kang2024few} integrates Image-Based Rendering (IBR) with NeRF to achieve superior performance in remote sensing applications. Additionally, depth smoothness constraints based on segmentation masks were employed to refine the geometry of remote sensing scenes. Lv et al. \cite{lv2023neural} introduced a novel NeRF-based method for high-resolution remote sensing view synthesis, incorporating attention mechanisms with neural radiance fields. Wu et al. \cite{wu2022remote} combined implicit multiplane image (MPI) representations with deep neural networks to model the 3D world, considering the overhead perspective and deep imaging characteristics of remote sensing images. Sat-NeRF \cite{mari2022sat} was designed to integrate contemporary neural rendering techniques with native satellite camera models, represented by rational polynomial coefficient (RPC) functions, allowing it to generate novel views from collections of satellite images captured from various viewpoints and dates. Earth Observation NeRF (EO-NeRF) \cite{mari2023multi} was specifically developed for Earth observation tasks, enabling the generation of novel satellite views from a set of multi-date satellite images.

\textbf{Applications on SAR: }
Several researchers have explored NeRF-based models by integrating synthetic aperture radar (SAR) imaging mechanisms with neural networks for SAR target generation. For example, Lei et al. \cite{lei2024sar} introduced a SAR-NeRF model that synthesizes SAR target images by leveraging the principles of neural radiance fields. Additionally, Radar Fields \cite{ehret2024radar} were developed to train a volumetric model using extensive SAR measurements of a scene, capturing its detailed 3D structure. Zhang et al. \cite{zhangCircularSARIncoherent2023} utilized a NeRF-like model for Circular SAR 3D inversion, enabling the reconstruction of target geometries from circular SAR data. NeRF has also been applied to generate novel view inverse synthetic aperture radar (ISAR) images. For instance, RaNeRF \cite{liu2023ranerf} combines NeRF with ISAR imaging models to represent the 3D structure of a target as a continuous 6D function, accounting for both spatial positions and viewing orientations. This model establishes a connection between the 3D structure and 2D ISAR images, facilitating the differential rendering of ISAR images. Another NeRF-based method, ISAR-NeRF \cite{deng2024isar}, encodes the 3D physical geometry of the target into an implicit neural network (INN) to infer novel 2D view images with consistent geometric fidelity.

\begin{table*}[h!]
    \centering
    \caption{The corresponding literature of computer vision and SAR domain that addressed the general challenges are summarized.}
    \label{tab:challenges}
    \begin{tabular}{ccccccccccccc}
    \toprule
    \textbf{\makecell[c]{Controllable \\ Generation}}
    & &\multicolumn{4}{c}{\textbf{\makecell[c]{cGAN-based \\ Control}}}
    &\multicolumn{3}{c}{\textbf{\makecell[c]{Implicit \\ Control}}}
    &\multicolumn{3}{c}{\textbf{\makecell[c]{Explicit \\ Control}}} &\textbf{\makecell[c]{Conditional \\ Diffusion Models}} \\
    \midrule
    \textbf{SAR} 
    & &\multicolumn{4}{c}{\makecell[c]{\cite{guoSARImageData2022,juSARImageGeneration2024,huangHighResolutionSAR2019,kongSARTargetRecognition2021,duHighQualityMulticategorySAR2022,zhuLIMEBasedDataSelection2022,caoDemandDrivenSARTarget2022,shiISAGANHighFidelityFullAzimuth2022,juSARImageGeneration2024}\\ \cite{zouMWACGANGeneratingMultiscale2020,qinTargetSARImage2022,pengGenerationSARImages2022,sunAttributeGuidedGenerativeAdversarial2023,ohPeaceGANGANBasedMultiTask2021,caoLDGANSyntheticAperture2020,guoSyntheticApertureRadar2017,zouMWACGANGeneratingMultiscale2020,guoSARImageData2022}}}
    &\multicolumn{3}{c}{\cite{fengInterpretationLatentCodes2023a}}
    &\multicolumn{3}{c}{\cite{giry2022sar,guoCausalAdversarialAutoencoder2023,huFeatureLearningSAR2021}}
    &\cite{shiBraininspiredApproachSARtooptical2024,baiConditionalDiffusionSAR2024}\\
    \textbf{CV} 
    & &\multicolumn{4}{c}{\cite{miyato2018cgans}}
    &\multicolumn{3}{c}{\cite{shoshanGANControlExplicitlyControllable2021,ge2020zero}}
    &\multicolumn{3}{c}{\cite{harkonen2020ganspace,wuStyleSpaceAnalysisDisentangled2021,zhuLinkGANLinkingGAN2023}}
    &\cite{dhariwal2021diffusion,ho2022classifier,rombachHighResolutionImageSynthesis2022,zhangAddingConditionalControl2023}\\
    \midrule
    \midrule
    \textbf{\makecell[c]{Data Constraint \\ Generation}} 
    &\multicolumn{2}{c}{\textbf{Augmentation}} &\multicolumn{2}{c}{\textbf{Regularization}} &\multicolumn{2}{c}{\textbf{Model Design}} &\multicolumn{3}{c}{\textbf{\makecell[c]{Knowledge \\ Transfer}}} 
    &\multicolumn{3}{c}{\textbf{\makecell[c]{Latent \\ Code}}} \\
    \midrule
    \textbf{SAR}
    &\multicolumn{2}{c}{\cite{luo2020synthetic,songLearningGenerateSAR2022,guoSyntheticApertureRadar2017}} 
    &\multicolumn{2}{c}{\cite{juSARImageGeneration2024,wangSyntheticApertureRadar2019,marmanis2017artificial}} 
    &\multicolumn{2}{c}{\cite{zouMWACGANGeneratingMultiscale2020,guoSARImageData2022,shiISAGANHighFidelityFullAzimuth2022}} 
    &\multicolumn{3}{c}{\cite{sunAttributeGuidedGenerativeAdversarial2023}} &\multicolumn{3}{c}{\cite{huFeatureLearningSAR2021,guoCausalAdversarialAutoencoder2023}} \\
    \textbf{CV}
    &\multicolumn{2}{c}{\cite{zhao2020differentiable,karras2020training,jiang2021deceive}} 
    &\multicolumn{2}{c}{\cite{zhang2019consistency,zhaoImprovedConsistencyRegularization2021,tsengRegularizingGenerativeAdversarial2021,fang2022diggan,ojhaFewshotImageGeneration2021,xiaoFewShotGenerative2022,wu2024domain,zhaoCloserLookFewshot2022,zhang2022towards}} 
    &\multicolumn{2}{c}{\cite{liu2020towards,cuiGenCoGenerativeCotraining2022,saxenaReGANDataEfficientGANs2023}} 
    &\multicolumn{3}{c}{\cite{yangOneShotGenerativeDomain2023,alanovStyleDomainEfficientLightweight2023,zhaoExploringIncompatibleKnowledge2023,zhao2022few,phaphuangwittayakulFastAdaptiveMetaLearning2022}} &\multicolumn{3}{c}{\cite{zhengWhereMySpot2023,dingAttributeGroupEditing2022,hong2022deltagan,xieLearningMemorizeFeature2022}} \\
    \midrule
    \midrule
    \textbf{\makecell[c]{Prior Guided \\ Generation}}
    &\multicolumn{5}{c}{\textbf{Learned Prior}}
    &\multicolumn{7}{c}{\textbf{Explainable Prior}} \\
    \midrule
    \textbf{SAR}
    &\multicolumn{5}{c}{\cite{niuImageTranslationHighresolution2021,wangSARtoOpticalImageTranslation2022}}
    &\multicolumn{7}{c}{\cite{zhangComparativeAnalysisEdge2021,liMultiscaleGenerativeAdversarial2022,weiCFRWDGANSARtoOpticalImage2023,guoEdgePreservingConvolutionalGenerative2021,zhangSARtoOpticalImageTranslation2024,leeCFCASETCoarsetoFineContextAware2023}} \\
    \textbf{CV}
    &\multicolumn{5}{c}{\cite{xuMultiViewFaceSynthesis2021,lahiriPriorGuidedGAN2020,yangGANPriorEmbedded2021,zhangPetsGANRethinkingPriors2022,yu2021pixelnerf,zhang2022nerfusion,muller2022autorf,johari2022geonerf,trevithick2021grf,jain2021putting}}
    &\multicolumn{7}{c}{\cite{darPriorGuidedImageReconstruction2020,zhangPetsGANRethinkingPriors2022,weiLiDeNeRFNeuralRadiance2024a,truongSPARFNeuralRadiance2023,deng2022depth,roessleDenseDepthPriors2022,liuNeuralRaysOcclusionaware2022,yangFreeNeRFImprovingFewShot2023,kimInfoNeRFRayEntropy2022,niemeyerRegNeRFRegularizingNeural2022}} \\
    \bottomrule
    \end{tabular}
\end{table*}

\subsection{Controllable Generation Methods}

Controllable generation has been attracting most attentions in SAR domain, such as simulating SAR images under targeted physical properties and imaging parameters. In computer vision domain, Wang et al. \cite{wangControllableDataGeneration2024} proposed to review the controllable deep data generation in depth. 

In this section, we will briefly summarize the current controllable GAN approaches in computer vision domain, targeting similar issues with SAR applications.

\subsubsection{cGAN-Based Methods}

The preliminary approaches came from conditional GAN model, where the generator and discriminator are conditioned on some additional conditions such as class labels \cite{mirza2014conditional}. The condition signals are concatenated with the input either at the input layer or at the hidden layer \cite{reed2016generative}. Multiple conditions can be concatenated together. However, the traditional cGAN is limited in precise control with complex conditional signals. In order to improve the generation quality and stablize the training, ACGAN was proposed that provided a stable supervision signal additionally \cite{odena2017conditional}. Miyato et al. \cite{miyato2018cgans} indicated that the direct concatenation of conditions is arbitrary and may cause expand the hypothesis space of the discriminator, causing the discriminator to be optimized towards certain functions that lack a logical mathematical foundation. To this end, the authors proposed a projection-based method for integrating conditional information into the discriminator of GANs, which accurately accounts for the function of the conditional information in the underlying probabilistic model \cite{miyato2018cgans}.

In 2018, the StyleGAN \cite{karras2019style} was first proposed that introduced a novel latent space representation, which is learned hierarchically with each level representing different attributes of the image. The so-called $\mathcal{W}$ space enables the generation of images with fine-grained control over various attributes. From then on, the controllable GAN entered into latent space optimization based approaches. It can be categorized into implicit and explicit control as demonstrated below.

\subsubsection{Implicit Control Based Methods}

The latent space representation of GAN can be disentangled so that the modification over the disentanglement features can control the generation. However, these methods can only alter the relative strength of specific characteristics, but they do not have the ability to directly establish their values. As a result, we conclude them as implicit control based methods. Among them, the GAN inversion has attracted the most attention.

GAN inversion aims to recover the input latent vector from a given generated image using a trained GAN model \cite{xia2022gan}. It can be regarded as an inversion process of generation. Specifically, GAN inversion provides a mechanism for elucidating the GAN’s latent space and delineating the process by which the images are generated, thus revealing the inner workings of deep generative models. The problem definition of GAN inversion can be expressed as follows. Denoting the image to be inverted as $x$, the pre-trained generator as $G$, and the latent vector as $z$. The GAN inversion problem is defined as the following optimization problem \cite{xia2022gan}:
\begin{equation}
    z^* = \arg\min_z L(G(z),x),
\end{equation}
where $L$ denotes the distance.

As we know, the GAN inversion is not the destination. The inverted code of a real image can be varied for a certain attribute, so that the image with different properties can be interpretably manipulated. Some selected GAN inversion based methods are introduced as follows. Härkönen et al. \cite{harkonen2020ganspace} discovered significant latent directions for producing interpretable controls for image synthesis using Principal Component Analysis (PCA) in either the latent space or feature space. The principal components acquired by PCA corresponded to certain attributes, and the interpretable controls of various image attributes can be defined by perturbing each layer along the principal directions. Similar to \cite{harkonen2020ganspace}, some researches conducted GAN inversion in $\mathcal{W}$ space. However, when multiple attributes are involved, modifying one value may have an impact on another property due to the lack of separation in certain semantics. To this end, Wu et al. \cite{wuStyleSpaceAnalysisDisentangled2021} suggested using the StyleSpace $\mathcal{S}$, which is the space defined by the style parameters for each channel. Empirical studies demonstrated that the StyleSpace has the ability to mitigate spatially intertwined alterations and apply accurate localized adjustments. By directly modifying the style code, this method can manipulate multiple properties and semantic directions without causing any impact on others. Although some subspaces of the latent space can influence specific semantics over generated images, there was no clear link between the local areas and the stated axes of latent spaces. In order to achieve this objective, Zhu et al. \cite{zhuLinkGANLinkingGAN2023} introduced a regularizer called LinkGAN, which explicitly connects the axes to arbitrary divisions of synthesized images. This technique allowed for precise local control of GAN generation by resampling partial latent codes.



\subsubsection{Explicit Control Based Methods}

Note that the implicit control can only reveal which latent code is responsible for the specific image attributes. In contrast, the explicit control further enables better interpretation of the latent codes. By employing contrastive learning, Xie et al. \cite{shoshanGANControlExplicitlyControllable2021} obtaind GANs that possessed a clearly separated latent space. This disentanglement technique was employed to train control-encoders that map human-interpretable inputs to appropriate latent vectors, enabling explicit control. Motivated by the visual generalization of primates, Ge et al. \cite{ge2020zero} provided a new learning framework called Group-Supervised Learning (GSL), which involves a family of objective functions applied to groups of examples. The training samples in each group have interpretable and interchangeable attributes. The GSL framework can learn to disentangle the inputs into multiple interchangeable components by explicitly swapping them across training samples, which may then be combined to generate novel examples.

\subsubsection{Conditional Diffusion Models}
In order to enable the diffusion models to control the generation results according to specific requirements, conditional diffusion models were explored in different applications with different forms of conditions. Dhariwal et al. \cite{dhariwal2021diffusion} proposed to improve the generation ability of diffusion models with classifier guidance. An extra classifier was trained and gradients can be taken from it to guide diffusion models to generate images with specific labels. Later, Ho et al. \cite{ho2022classifier} introduced a classifier-free diffusion guidance, where a conditional and an unconditional diffusion model were trained jointly, making it possible to attain a trade-off between sample quality and diversity under the guidance. Besides labels, diffusion models can also be directed by information from other modal, such as texts, images and semantic maps. Latent Diffusion Models(LDM) \cite{rombachHighResolutionImageSynthesis2022} unified these conditions with flexible latent diffusion through a specially designed general-purpose conditioning mechanism based on cross-attention. To enable finer grained spatial control by provided additional images(e.g., edge maps, human pose skeletons, segmentation maps, depth, normals, etc.) that directly specify desired image composition, ControlNet \cite{zhangAddingConditionalControl2023} was proposed, treating the large pretrained model as a strong backbone for learning diverse conditional controls. Based on Stable Diffusion, by maintaining the quality and capabilities of the large model by locking its parameters and makeing a trainable copy of its encoding layers, ControlNet was able to attain increased precision in spatial control.

\subsubsection{Applications on SAR}

Aiming to solve the controllable generation problem, the current applications for SAR can also summarized into cGAN, implicit control, and explicit control-based categories. Among them, the cGAN-based methods account for the main part.

The cGAN-based controllable SAR image generation methods follow a similar protocol, i.e., the condition factors (mainly including target label, angle, target location) are encoded as input, and multiple auxiliary classifiers are designed to evaluate the properties of generated images with input conditions. The strategies of condition embedding could be various. One of them is to encode the condition factor with noise vector in the input layer \cite{shiISAGANHighFidelityFullAzimuth2022,qinTargetSARImage2022,caoLDGANSyntheticAperture2020,caoDemandDrivenSARTarget2022,zhuLIMEBasedDataSelection2022,zouMWACGANGeneratingMultiscale2020,kongSARTargetRecognition2021,pengGenerationSARImages2022,sunAttributeGuidedGenerativeAdversarial2023,ohPeaceGANGANBasedMultiTask2021,guoCausalAdversarialAutoencoder2023}. Another one is to apply a projection module on the condition and multiply with features in the latent space \cite{shiISAGANHighFidelityFullAzimuth2022,duHighQualityMulticategorySAR2022}. The last one is to concatenate the condition with multiscale features \cite{songSARImageRepresentation2019,songLearningGenerateSAR2022}. Most studies explored the vehicle target generation with different categories and azimuth angles.

Some researches aim to address the data limited issue in SAR ship detection by image generation. Ju et al. \cite{juSARImageGeneration2024} and Guo et al. \cite{guoSARImageData2022} proposed a cGAN-based model to generate SAR ship images with constraint maps. The instance segmentation map was applied in \cite{guoSARImageData2022}. In contrast, the target location map based on bounding box together with a pixel value constraint map are applied as the input conditions in \cite{juSARImageGeneration2024}. Similarly, Zou et al. \cite{zouMWACGANGeneratingMultiscale2020} explored a multiscale wasserstein ACGAN to generate different types of SAR ship target images based on label condition. The generated SAR ship images are mixed with real data to train the detection model and tested on large scene SAR images.

The implicit control of SAR image generation refers to the methods where the latent space can be interpreted by target properties and new image can be generated by manipulating the latent codes, but the certain values of property cannot be obtained. Feng et al. \cite{fengInterpretationLatentCodes2023a} proposed an InfoGAN based approach to implicitly control the properties of "rotation, scaling, and shift" of SAR target. The relation between different properties of SAR images and the latent codes in the InfoGAN was preliminarily explained. However, this work also presented that different properties (more than 2) will lead to entanglement among latent codes, that is, each individual property cannot be controlled separately by latent codes. 

As a comparison, the explicit control enables assigning values to properties which are mapped to the latent code further to control the generation. The following two work are based on StyleGAN \cite{giry2022sar,guoCausalAdversarialAutoencoder2023}. Specifically, Guo et al. \cite{guoCausalAdversarialAutoencoder2023} proposed a causal model of SAR image generation and disentangled the semantic factors of SAR target in $\mathcal{W}$ space of StyleGAN, including intrinsic, diversity, and randomness. Hu et al. \cite{huFeatureLearningSAR2021} built an interpretable feature space based on the proposed CVAE-GAN model, where the features of SAR images are continuous and can represent some properties with interpretability to some degree. It was achieved by the continuous latent variable space in variational auto-encoder. To this end, the angular margin clearly appeared in the latent feature space, from which the inter-class discrimination and the intra-class similarity are improved. The learned feature space makes the model capable of recognizing the unknown classes by rejection.

Conditioning on the input SAR images, the conditional diffusion models allowed for the generation of corresponding optical images. The novel conditional diffusion model proposed by Bai et al. \cite{baiConditionalDiffusionSAR2024} for SAR-to-optical image translation enabled the preservation of the target information. In order to improve the ability to extract feature, 
self-attention and long-skip connection mechanisms were integrated with the noise prediction network where SAR images were used as conditional inputs \cite{shiBraininspiredApproachSARtooptical2024}.

\subsection{Data Constraint Generation Methods}

SAR image generation often occurs with problem of data constraint. In this section, we review the data constraint challenge and solutions in the literature, and then summarize the related researches in SAR image applications.

Limited data training will result in mode collapse where the generator fails to capture the full diversity of the data distribution. On the other hand, overfitting particularly happens in discriminator to only distinguish the specific examples in the training set and unable to generalize well to new and unseen data. GAN training is already known for its instability, and limited data can make the training process even more unpredictable. The solutions can be summarized in the following aspects.

\subsubsection{Augmentation}

Facing the challenge of limited training samples, the most intuitive solution is data augmentation. The traditional data augmentation methods include geometric transformation, color transformation, etc, which are widely applied for real or fake images or both of them to optimize the discriminator. However, the augmentation would lead to "leaking" problem that impels GAN to learn the augmented distribution. To solve this, Karras et al. \cite{karras2020training} proposed an adaptive discriminator augmentation mechanism where a diverse set of augmentations and an adaptive control scheme were designed to stabilize training significantly. If the real and fake images are both augmented when training the discriminator, a delicate equilibrium between the generator and the discriminator would be disrupted. To this end, the Differentiable Augmentation (DiffAugment) strategy was proposed \cite{zhao2020differentiable} that applied the same differentiable augmentation for both generator and discriminator training, effectively stabilizing the training and leading to better convergence. Apart from standard data augmentation methods, Jiang et al. \cite{jiang2021deceive} proposed an adaptive pseudo augmentation (APA) approach that employed the generator to augment the real data distribution. In this way, the discriminator was deceived adaptively by generator, and thus the problem of discriminator overfitting was mitigated.

\subsubsection{Regularization}

Limited training data directly results in discriminator overfitting. In machine learning community, model regularization has been a most popular strategy to alleviate overfitting. Consequently, there are a series studies that focus on optimizing discriminator by designing useful regularization terms. Consistency regularization, a popular strategy in semi-supervised learning  \cite{sajjadi2016regularization,laine2016temporal}, was introduced in GAN to stabilize discriminator training \cite{zhang2019consistency,zhaoImprovedConsistencyRegularization2021,tsengRegularizingGenerativeAdversarial2021}. It is often applied together with data augmentation methods. The core idea is to penalize the sensitivity of the discriminator to the arbitrary semantic-preserving perturbations from data augmentation \cite{zhang2019consistency}. Subsequently, Zhao et al. \cite{zhaoImprovedConsistencyRegularization2021} proposed the improved consistency regularization, combining balanced consistency regularization (bCR), latent consistency regularization (zCR), and CR, for generated images, the latent vector space, and the generator, respectively. 

Some other literature target more rigorous limited data generation task, where only 10\% or 20\% of CIFAR-10 or CIFAR-100 dataset were applied for training in the experiments. They can be integrated to arbitrary GAN models with or without data augmentation. A novel regularization term was proposed in \cite{tsengRegularizingGenerativeAdversarial2021}, which recorded the L2 norm between the current prediction of the real image and a moving average result of historical predictions of the generated image, in order to modulate the discriminator's prediction of learning a robust GAN model. DigGAN, as known as Discriminator gradIent Gap, was introduced \cite{fang2022diggan} to narrow the gap between the norm of the gradient of a discriminator's prediction with respect to real images and generated data.

In few-shot generative model adaption task, the main challenges lie in cross-domain inconsistency and poor diversity in target domain image generation. There are some regularization approaches were proposed, to either facilitate the structure information alignment in cross-domain or improve the diversity of generation. To enhance cross-domain consistency, Wu et al. \cite{wu2024domain} proposed a structure loss based on similarity, which facilitates the alignment of the auto-correlation map of the target image with that of the source image throughout the training process. To align spatial structural information between source and target domain images, Xiao et al. \cite{xiaoFewShotGenerative2022} proposed Relaxed Spatial Structural Alignment (RSSA), that preserves and transfers source domain structure and spatial variations to the target domain by compressing the latent space to a target-domain proximate subspace and regularizing self-correlation and disturbance correlation consistencies. A novel one-shot generative domain adaptation method named DiFa was designed in \cite{zhang2022towards}. To promote variability in generation, the authors employed a selective cross-domain consistency mechanism, which specifically identifies and preserves domain-sharing attributes within the editing latent space W+. Zhao et al. \cite{zhaoCloserLookFewshot2022} proposed to apply mutual information (MI) maximization achieved by contrastive loss (CL) to prevent the generator from overfitting, which retains the source domain’s multi-level diversity information in the target domain generator. Similarly, Ojha et al \cite{ojhaFewshotImageGeneration2021} presented strategies for transferring diversity from a large source domain to a target domain, including a cross-domain consistency regularization to maintain pairwise distances and an anchor-based approach to enhance synthesis fidelity and reduce overfitting in the latent space.

\subsubsection{Model Design} 

To alleviate the model overfitting, some studies aim to improve model architectures. A light-weight GAN structure called FastGAN was introduced by Liu et al. \cite{liu2020towards}. It was made up of a self-supervised discriminator with an additional decoder that optimized it for more descriptive feature-maps and richer training signals for the generator, and a skip-layer channel-wise excitation module that improved gradient flow for faster training. Some researches applied knowledge distillation or pruning to reduce the model parameters. For instance, Saxena et al. \cite{saxenaReGANDataEfficientGANs2023} proposed Re-GAN, a data-efficient GAN training approach that adapts the GAN architecture dynamically during training, investigating various sub-network configurations in real-time. In order to regularize the GANs network and lower the possibility of prematurely pruning crucial connections, Re-GAN periodically prunes irrelevant connections and regrows them. Additionally, Cui et al. \cite{cuiGenCoGenerativeCotraining2022} proposed GenCo, a Generative Co-training network, which addresses discriminator overfitting by incorporating multiple complementary discriminators offering diverse supervision from unique perspectives during training.

\subsubsection{Knowledge Transfer}

Some researches focus on improving the knowledge transfer with limited data, especially for few-shot generative model adaption task. One of the efficient strategies to improve knowledge transfer is reducing the trainable parameters and maintaining the quality. To this end, Yang et al. \cite{yangOneShotGenerativeDomain2023} proposed an attribute adaptor in the generator. In this way, only one layer in the generator and discriminator need to be trained and the prior knowledge is reused to the most extent. In addition, the quality and diversity of generated images are maintained well. Based on StyleGAN, Alanov et al. \cite{alanovStyleDomainEfficientLightweight2023} proposed  two novel parameterizations for few-shot domain adaptation, StyleSpaceSparse and Affine+, which reduce the number of trainable parameters and maintain quality. StyleSpaceSparse further minimizes parameters while AffineLight+ optimizes with significantly fewer parameters than training StyleGAN, achieving state-of-the-art performance for few-shot adaptation to dissimilar domains. On the other hand, some researches aim to address the challenge of incompatible knowledge transfer. Zhao et al. \cite{zhaoExploringIncompatibleKnowledge2023} proposed a knowledge truncation method, so called Removing In-Compatible Knowledge (RICK) based on lighweight filter-pruning to solve this. It removes filters that encode incompatible knowledge during FSIG adaptation. Zhao et al. \cite{zhao2022few} also addressed the knowledge preserving issue where only source domain and source task are considered while ignoring the target information. The Adaptation-Aware kernel Modulation (AdAM) method was proposed to improve the few-shot generation of different source-target domain proximity. At last, the meta-learning was also introduced into generative models. Phaphuangwittayakul et al. \cite{phaphuangwittayakulFastAdaptiveMetaLearning2022} proposed the Fast Adaptive Meta-Learning (FAML) optimization method for GAN.

\subsubsection{Latent Space Optimizing}

The latent code of the unseen classes is lack of diversity, which leads to poor generation performance. The previous studies demonstrated that the latent space in GAN could be distangled and interpretable for different attributes. To this end, given a pre-trained GAN model, it is possible to generate images of unseen class with combinations of attributes from seen classes. Yet, a set of studies proposed to optimize the latent space to represent the unseen classes with more diversity. For example, Zheng et al. \cite{zhengWhereMySpot2023} investigated the consistency of the latent space in order to identify previously unseen categories. The rationale was that the adjacent latent space around the new class belongs to the same category. The optimization contains two stages, with which the latent space of previously unseen categories can be further improved, resulting in an enhanced ability to generate latent codes. Some other studies aim to distangle the class-relevant and class-independent representations in the latent space, so as to generate diverse images by modifying the class-independent codes based on a few samples. Ding et al. \cite{dingAttributeGroupEditing2022} developed an inversion-based approach in attribute group editing (AGE) which used class embeddings to represent the qualities and learned a dictionary to gather the attributes irrelevant to the category. Thus, the class embeddings and dictionary allowed for the generation of various images by modifying the category-irrelevant properties while preserving category-relevant attributes inside the latent space. Hong et al. \cite{hong2022deltagan} proposed a Delta Generative Adversarial Network (DeltaGAN) which consists of a reconstruction and a generation subnetwork. The former aims to extract the intra-category information referred to as "delta". The latter produced a sample-specific "delta" that is subsequently combined with the input image to generate a new image belonging to the same category. Xie et al. \cite{xieLearningMemorizeFeature2022} introduced FeaHa where the category-specific and category-agnostic features are separated. The category-agnostic features were used to construct a memory bank of reusable features using a feature hallucination module. Then, it is possible to generate diverse images by sampling from the memory bank.

\subsubsection{Applications on SAR}

The data constraint issue particularly received attention in SAR image generation due to the high cost of data acquirement. We summarize the related work that addressed this issue corresponding to some representative literature in computer vision domain, as given in Table \ref{tab:challenges}.

Regarding data augmentation, Luo et al. \cite{luo2020synthetic} applied a simple over-sampling data augmentation method for minority class by searching a best augmentation policy automatically. Some studies discovered the intrinsic data problems of SAR image that would cause mode collapse and overfitting of GAN model. Guo \cite{guoSyntheticApertureRadar2017} indicated that the distinguished clutter level in SAR image training set would influence the discriminator's decision. Consequently, a clutter normalization method was proposed to prevent the discriminator distinguishing the real and fake images according to clutters. Song et al. \cite{songLearningGenerateSAR2022} demonstrated the problem of conventional rotation representation which is not linearly accumulative. It would require a large number of nonlinear neurons to learn so as to cause overfitting with limited training samples under sparse observations. To solve this, the authors proposed a rotated cropping method for rotation representation.

The traditional regularization terms of GAN are also applied in SAR domain. For example, the multi-scale feature distance based loss of generator and the perceptual loss were introduced to improve the structure and semantic information in \cite{juSARImageGeneration2024}. Besides, some studies designed the regularization terms according to the specific characteristics of SAR image. Marmanis et al. \cite{marmanis2017artificial} first proposed a histogram-based loss to constrain the global intensity distributions of the generated SAR images, which improved the generation performance to a certain extent. Note that the proposed method is conducted on TerraSAR-X high-resolution satellite scene images with complex textures which is different from most generation method for SAR targets. Wang et al. \cite{wangSyntheticApertureRadar2019} proposed a reconstruction loss based on Fisher-Tippet distribution to describe the speckle statistics, which led to the generated images with lower speckle corruption.

The improvement of model architecture design has also been explored in SAR image applications. Zou et al. \cite{zouMWACGANGeneratingMultiscale2020} applied the multi-scale discriminator from low to high resolution, offering multiple complementary discrimination losses during training. Guo et al. \cite{guoSARImageData2022} introduced the residual and attention block to prevent the generator from gradient vanishing. Shi et al. \cite{shiISAGANHighFidelityFullAzimuth2022} proposed a lightweighted simplified self-attention GAN to reduce the model parameter. 

Only a few studies focused on the few-shot image generation or few-shot generative model adaption task in SAR domain. Sun et al.  \cite{sunAttributeGuidedGenerativeAdversarial2023} explored two few-shot cases of SAR image generation for the first time. A meta-learning based improved episode training strategy was proposed in this work, where the episodes were constructed by source and target data with specified class label definition. In this way, the angle information learned from source domain is transferred to target domain even they do not share the same semantic labels. 

The following studies proposed by Hu et al. \cite{huFeatureLearningSAR2021} and Guo et al. \cite{guoCausalAdversarialAutoencoder2023} adopted the idea of optimizing the latent code representation to improve the diversity and quality of the generated image. For comparison, Hu et al. \cite{huFeatureLearningSAR2021} proposed to optimize the latent space of image so that the latent codes can express the semantic meanings. Guo et al. \cite{guoCausalAdversarialAutoencoder2023} introduced the diversity and randomness into the latent space as a complementary of intrinsic properties of SAR targets, so that the generalization of the latent codes were improved a lot.

\subsection{Prior Guided Generation Methods}

In this paper, two different priors are defined according to their representation. 

\subsubsection{Learned Prior}

The learned prior is defined as the prior information learned from data and it may be not explainable, such as the feature representation, semantic information of a pre-trained deep model, latent codes from GAN inversion, etc.

Some selected work based on GAN model are introduced as follows. Xu et al. \cite{xuMultiViewFaceSynthesis2021} proposed Face Flow-guided Generative Adversarial Network (FFlowGAN), which designed a face flow module to learn a dense correspondence between the input and target faces, offering strong instruction for the face synthesis module to emphasize significant facial texture. Lahiri et al. \cite{lahiriPriorGuidedGAN2020} proposed to learn a data driven parametric network to directly predict a matching prior for a given masked image, which was then utilized to reconstruct the signal given a well-pretrained generative model. In \cite{yangGANPriorEmbedded2021}, a GAN prior network was first trained, allowing for high-quality image generation and then embedded into the GAN prior embedded network (GPEN).The latent coding and noise inputs of the GAN network were substituted with the output of the fine-tuned DNN, which mapped blind face images to GAN priors, in order to guide the restoration process. In \cite{zhangPetsGANRethinkingPriors2022}, the authors proposed an external and an internal prior for single image generation, of which the external prior can be regarded as learned prior since it is obtained through GAN inversion. The strategy enhanced the generation of natural, coherent, and diverse samples.

The semantic information extracted from pre-trained models can be applied as priors in NeRF training. A CLIP ViT model trained on a large dataset of images and their corresponding natural language descriptions was applied in \cite{jain2021putting} as the learned prior. Based on this, the semantic consistency loss was proposed to encourage the rendered images to match the high-level semantic features of the input images across different poses. In AutoRF \cite{muller2022autorf}, the author suggested acquiring a standardized, object-focused representation that captures and separates shape, appearance, and position attributes. Concretely, a 3D object detection algorithm was employed alongside panoptic segmentation to extract 3D bounding boxes and object masks as 3D object priors. The priors offered highly generalized and concise information which was to the interest of the object, which was then decoded into a radiance field and utilized for novel view synthesis.

Besides, the pre-trained models also enrich the feature representations in terms of improving the generated model training. In pixelNeRF \cite{yu2021pixelnerf}, Yu et al. employed the pre-trained CNN to extract image information as a scene prior, which was further input into the NeRF along with positional encoding and the view direction. NeRFusion \cite{zhang2022nerfusion} proposed by Zhang et al. first created a local feature volume retrieved by a CNN which were combined into a global feature volume using a 3D CNN and then processed further with gated recurrent units (GRU). Ultimately, the global feature volume is used to generate the rendered output. The authors pointed out that the data priors provided suitable initial radiance fields and aided the reconstruction.


\subsubsection{Explainable Prior}
\label{sec:expprior}

The explainable prior is defined as the prior information derived from the explainable model or physical laws, such as edge information extracted with wavelet model, depth information obtained from LiDAR data, statistical distribution of a image, etc. 

Dar et al. \cite{darPriorGuidedImageReconstruction2020} introduced three kinds of priors into cGAN for image reconstruction, which provided high-spatial-frequency details, low-frequency constraint, and high-level semantic features. The integration of these priors within the cGAN framework offers a synergistic approach to enhance the quality and fidelity of the reconstructed images. In \cite{zhangPetsGANRethinkingPriors2022}, the authors proposed an external and an internal prior for single image generation, of which the internal prior can be considered as explainable prior since it reflected the patch distribution correction. The injection of internal priors helped to achieve fast restoration from low resolution to high resolution syntheses.

Specifically, the explainable priors are commonly introduced in NeRF especially for sparse view generation (what we deeply concern for SAR applications), including depth information and geometry information. Based on those prior information, some regularization loss terms were designed to constrain the model training. To tackle the geometric inaccuracy issue in NeRF reconstruction dealing with sparse input views, Wei et al. \cite{weiLiDeNeRFNeuralRadiance2024a} proposed LiDeNeRF by integrating the LiDAR-derived depth priors to enhance the accuracy of depth estimation and overall 3D reconstruction quality. The sparse depth prior was obtained by projecting LiDAR point clouds onto the image plane. A depth correction module was introduced to correct the depth estimation errors in specific ray directions, thus improving the convergence and accuracy of the predicted images. Similarly, Truong et al. \cite{truongSPARFNeuralRadiance2023} proposed a depth consistency loss to enhance the quality of the reconstructed scene by ensuring depth consistency from any viewpoint. The depth information of unknown view-points, different from the one in \cite{weiLiDeNeRFNeuralRadiance2024a}, was derived from geometric warping of training data. The depth information can be also obtained from Structure-from-Motion (SfM) preprocessing \cite{deng2022depth,roessleDenseDepthPriors2022}. In \cite{deng2022depth}, DS-NeRF was proposed that introduced a loss function to encourage the rendered rays’ termination distribution to match the depth provided by the 3D keypoints from SFM. It was compatible with other NeRF-based methods and can be integrated into existing pipelines without requiring additional annotations or assumptions. The dense depth maps along with their uncertainty estimates, were used to impose constraints during the NeRF optimization process, as illustrated in \cite{roessleDenseDepthPriors2022}.

The geometry prior, on the other hand, is also considered in some literature. Yang et al. \cite{yangFreeNeRFImprovingFewShot2023} indicated that the high-frequency components in positional encoding which reflected the object details of edges mainly cause NeRF to overfit under sparse view training. Therefore, a frequency regularization strategy was proposed where the high-frequency components were occluded during the early stage of training. In order to address the floating artifacts issue, Nie et al. \cite{niemeyerRegNeRFRegularizingNeural2022} proposed a geometry regularization to apply a smoothness loss on the rendered depth patches. In this way, the geometry inconsistency among different views was alleviated. 

\subsubsection{Applications on SAR}

Many related literature leverage prior guided generation strategies for SAR-to-Optical translation. Niu et al. \cite{niuImageTranslationHighresolution2021} applied a pre-trained VGG-16 model to extract the features from real optical images, denoted as guided features. They were then utilized to effectively guide the training of generator. Wang et al. \cite{wangSARtoOpticalImageTranslation2022} designed an optical image reconstruction sub-network parallel with the SAR-to-Optical sub-network and constrained the intermediate features within the two generators. The optical image reconstruction sub-network offered feature priors of optical data that guided the translation model for training. 

Besides the learned priors from optical images, there are also some explicit priors obtained with explainable models. Thermodynamics indicates the molecules will achieve a stable state in a heat field, which inspired the design of the first law of thermodynamics (FLT)-guided branch to regularize the feature diffusion and preserve image structures during S2O image translation \cite{zhangSARtoOpticalImageTranslation2024}. Lee et al. \cite{leeCFCASETCoarsetoFineContextAware2023} introduced the NIR image into the SAR-to-EO(RGB) translation method as an explicit data prior, where the SAR-to-NIR translation task was constructed to assist the main task to learn discriminative characteristics of various SAR objects. Edge information is a popular prior for SAR-to-Optical translation task which were applied in many literature \cite{zhangComparativeAnalysisEdge2021,liMultiscaleGenerativeAdversarial2022,guoEdgePreservingConvolutionalGenerative2021}. The common strategy is to fuse the edge feature with deep features in the generator to enrich the representation. The wavelet priors are also explored in \cite{weiCFRWDGANSARtoOpticalImage2023,liMultiscaleGenerativeAdversarial2022} in terms of fusion. Li et al. \cite{liMultiscaleGenerativeAdversarial2022} proposed to learn a mapping from SAR images to wavelet features with the generator, and the discriminator was designed to distinguish not only the gray-scale image but also the wavelet features. Some other priors were also discussed in terms of the effectiveness for generation task and detection task, such as polarimetric features \cite{zhangComparativeAnalysisEdge2021,gao2023dualistic,gao2023scattering}.


\section{Interpretable Model Based Simulation}

The traditional SAR image simulation methods are basically relied on interpretable physical models. According to different theoretical foundations, the interpretable physical models can be categorized into electromagnetic (EM) model based on Maxwell's equations, and the statistical model based on SAR image itself. Consequently, the SAR image simulation can be achieved either by EM simulation or statistical model based numerical calculation. In this section, we will briefly introduce the interpretable model based SAR image simulation approaches and the physics behind, since they play an important role for physics-guided hybrid modeling towards developing trustworthy GenAI algorithms. 


\subsection{EM Simulation}

EM model aims to simulate the interaction between electromagnetic waves and the surface of an object and to calculate the scattering field. It typically involves the use of Maxwell's equations and boundary conditions, combined with numerical methods such as integral equations and difference equations, to obtain an accurate distribution of the scattering field. Some typical physical theories involved in the EM models include physical optics (PO) \cite{asvestas1980physical}, geometrical optics (GO) \cite{greivenkamp2004field}, geometrical theory of diffraction (GTD) \cite{keller1962geometrical}, ray tracing \cite{auerRayTracingSimulationTechniques2010}, shooting and bouncing ray (SBR) \cite{kasdorfAdvancingAccuracyShooting2021}, mapping and projection algorithm \cite{xuImagingSimulationPolarimetric2006}. To this end, it is essential to obtain the CAD model and the surface parameters of the object, as well as the imaging parameters of sensor, so as to input them to the simulation system. Based on the EM model, the simulation can be realized in two ways, including SAR image simulation which provides the focused SAR image directly and raw echo simulation which considers the SAR system processing \cite{franceschettiSARASSyntheticAperture1992,gelautzSARImageSimulation1998,changSARIMAGESIMULATION2011}. 

Aiming at SAR image simulation for applications in urban areas, Balz et al. \cite{balzPotentialsLimitationsSAR2015} summarized and analyzed the current popular SAR image simulator, i.e., RaySAR \cite{auerRayTracingSimulationTechniques2010,auerRaySAR3DSAR2016} based on SBR, CohRaS \cite{hammer2009coherent}, and SARViz \cite{balzHybridGPUBasedSingle2009}, which can simulate the focused SAR image directly. Other SAR image simulator include MOCEM \cite{cochin2008mocem} and Xpatch \cite{researchandtechnologyorganizationNoncooperativeAirTarget1998}. Some public simulated datasets produced by them, such as Synthetic and Measured Paired Labeled Experiment (SAMPLE) \cite{lewis2019sar}, were open to public for researches. The raw signal simulation aims to calculate the reflectivity map based on EM model and simulate the raw signal in terms of the SAR system function. Then, the imaging processing is conducted to obtain the focused SAR image. Some representative studies include but not limit to \cite{franceschettiSARASSyntheticAperture1992,gelautzSARImageSimulation1998,changSARIMAGESIMULATION2011,franceschettiSARRawSignal2003}. Recently, SARCASTIC-v1.0 and SARCASTIC-v2.0 were proposed to simulate raw echo signal by generating phase history data for each pulse in the simulated collection \cite{woollardSARCASTICV2HighPerformance2022}.


\subsection{Statistical Model}

The statistical method does not consider the specific scattering and imaging mechanism of SAR. Instead, it only relies on the distributional properties of pixel values and their neighbors. Compared with EM model, it is more straightforward but lacks the same level of physical foundation. SAR image simulation based on statistical model can be briefly summarized as the following steps. 1) Constructing a statistical model to describe the distribution of the SAR image. 2) Estimating the parameters of the statistical model. 3) Obtaining simulated image based on the statistical model with estimated parameters. It has been widely applied for clutter simulation to generate various background and integrated with foreground.

Some typical statistical models are applied to describe different SAR clutters \cite{gao2017cfar,gao2016scheme,gao2018scheme}, e.g., the Rayleigh distribution for homogeneous areas \cite{oliver2004understanding}, the K distribution for intermediate inhomogeneous area \cite{jakeman1980statistics}, the G0 distribution for extremely inhomogeneous areas \cite{freryModelExtremelyHeterogeneous1997a}. For urban areas corresponding to heavy-tailed distributions, the $\alpha$-stable distribution was proposed as a generalization of the Rayleigh distribution \cite{kuruogluModelingSARImages2004}. Some other studies also investigated the statistical modeling for polarimatric SAR data \cite{dengStatisticalModelingPolarimetric2017,dengPhysicalAnalysisPolarimetric2016}. Yue et al. \cite{yueSyntheticApertureRadar2021,yueSyntheticApertureRadar2021a} conducted a thorough review regarding SAR image statistical modeling, including single-pixel statistical models and spatial correlation analysis. 

Based on the constructed statistical models, one can simulate SAR clutter images with same distribution. Collins et al. \cite{collinsModelingSimulationSAR2009} carried out the raw SAR data simulation based on the statistical characteristics of elemental scene scatterers using cSAR simulator. Yue et al. \cite{yueGeneralizedGaussianCoherent} proposed a simulation approach for correlated SAR texture based on the Gaussian coherent scatterer model. Another typical application for statistical modeling is sea surface simulation reported in literature  \cite{landyFacetBasedNumericalModel2019,rizaev2022modeling,liNumericalSimulationSAR2022a}.

\section{Hybrid Modeling}
\label{sec:hybridmodeling}

Generative AI models offer advantages in terms of cost, flexibility, and scalability, whereas they have limitations related to their high data requirements, inconsistencies in physics, limited ability to generalize, and lack of interpretability in SAR applications. In contrast, the physical model based simulation approaches consider the intrinsic physics foundation and empirical statistics of SAR which demonstrate interpretability and data efficiency. Nevertheless, they also have obstacles of obtaining accurate physical parameters of real-world, such as surface parameters and 3D model of objects. To this end, hybrid modeling refers to the integration of GenAI models and physics-based models in order to achieve mutual complementarity between both.

Such hybrid modeling techniques for SAR image generation can be categorized into AI-empowered physical modeling approaches and physics-inspired learning methods. The former one basically follows the physics based framework, where AI approaches substitute some sub-modules with flexibility, or build connections between forward and inverse process, to optimize the simulation. The latter one is constructed mainly based on data-driven GenAI framework, but the inner designs, such as data input, model architecture and loss function, are in accordance with physical laws.


\subsection{AI-Empowered Physical Modeling}

EM based simulation requires accurate physical parameters and 3D model of object as inputs to achieve high quality simulation results. Similarly, the numerical simulation for SAR clutters requires a proper model selection and accurate statistical parameter estimation. In real-world application scenarios, collecting the precise 3D model of target and measuring the accurate physical parameters are difficult that restrict the simulation performance. Recently, some deep learning based approaches were proposed to estimate the physical parameters or reconstruct the 3D model of objects from measured SAR images. In this way, the deep neural networks act as an sub-module within the physical simulation process in substitution of the former interpretable process. We refer to these approaches as AI-powered physical modeling in this paper.

Surface parameters of objects and scenes, including roughness, correlation length, permittivity, and geometry, are crucial for previously introduced SAR image simulators, especially in case of simulating a complex environment. In order to achieve more accurate and automatically adjustable parameter setting in EM simulation, Niu et al. \cite{niuParameterExtractionBased2020} considered a regression problem to estimate the simulation parameters from measured SAR images. A deep learning framework was built based on GAN and CNN, which was trained with simulated image from RaySAR engine as the input and the corresponding simulation parameters as labels. Thus, the trained model is embedded into the simulation engine to provide object parameters estimated from real images. The authors also proposed a DNN embedded simulation method that can calculate the reflection from measured SAR image with deep learning \cite{niuSARTargetImage2021}. Wang et al. \cite{wangMultiParameterInversionAIEM2022} proposed a bi-directional deep neural network based Advanced Integral Equation Model (AIEM) for surface parameter inversion (dielectric constant and roughness) from backscattering coefficient. It is constructed a forward and backward neural network to predict the backscattering coefficient and inverse the surface parameters, thus improving the performance by optimizing the forward and inverse processing interactively. Similar researches can be referred to EM imaging \cite{guoPhysicsEmbeddedMachineLearning2023} where EM parameters are solved given measured electric fields.



Another representative researches aim to develop differential processing to represent or solve the forward and inverse problem using deep learning. Song et al. \cite{songPhysicalawareRadarImage2021} proposed a projection network that explicitly represented the projection process of SAR imaging with a differential layer. It is trainable and can be integrated with other trainable layers to construct a explainable and hybrid model for SAR image generation. Thus, the SAR-specific layover effects can be realized in such model.

Differentiable rendering, on the other hand, is another typical technology for AI-powered physical modeling. The core idea is to model the rendering process as a differentiable function that takes rendering parameters as input and produces a rendered image as output. By computing the gradient of this function with respect to its input parameters, they can be used to iteratively optimize the rendering parameters, thus improving the quality of the rendered image. Note that the previous introduced ray tracing based SAR simulator, such as RaySAR, requires precise 3D model and other scene parameters to render the SAR image. The pre-defined parameters would result in inferior simulation results. Recently, a series of differentialable rendering based approaches were proposed to combine the forward and inverse problem for optimization.

Fu and Xu \cite{fuReconstruct3DGeometry2021} first proposed a differentiable SAR renderer primarily inspired by SoftRas \cite{liuSoftRasterizerDifferentiable2019}, that introduced the projection and mapping plane to derive the SAR image \cite{xuImagingSimulationPolarimetric2006}. It tentatively achieved the 3D reconstruction of T72 target. Subsequently, the differentiable SAR renderer (DSR) was proposed \cite{fuDifferentiableSARRenderer2022}. By reformulating the MPA of SAR imaging mechanism in the differentiable form and deriving the first-order gradients, the error of rendered image can be back-propagated to optimize the 3D model and scattering attributes of target. The primary DSR appeared limitations in reconstruction from shadow and distorted scattering. To solve this problem, the authors further proposed an extension version of DSR \cite{fuExtensionDifferentiableSAR2023} to model the illumination map and shadow map for reconstruction. Additionally, a new metric was proposed to evaluate the smoothness of the mesh surface to improve the reconstruction performance. Note that DSR mainly solves single scattering and geometric reconstruction, while being difficult to model multiple scattering due to the rasterization approximation. Accordingly, the differential ray tracing (DRT) simulator was proposed that aimed to learn spatially varying surface scattering parameters from SAR images \cite{wei2024learning}. Jia et al. proposed to apply DSR for multi-view target generation for data augmentation, that facilitated the SAR target recognition in case of limited measurements \cite{JiaRongHeKeWeiFenXuanRanDeSARDuoShiJiaoYangBenZengGuang2024a}.


\subsection{Physics-Inspired GenAI}


In practice, as the optimization process is extremely non-convex, it is challenging to accomplish the convergence of a data-driven generative model to global minima. Optimization processes that converge to local minima without constraints may result in models that have limited generalization ability or results that violate extant knowledge, including intuition and physics laws. The conventional GenAI models lack domain knowledge of SAR, leading to inefficient model training, inconsistency with electromagnetic properties, and limited generalization capacity. The physics-inspired GenAI models aim to utilize principles from physics to include domain expertise by establishing the "SAR physical layer" \cite{datcuExplainablePhysicsAwareTrustworthy2023,huangProgressPerspectivePhysically2022} within the GenAI models, thus achieving faster optimization, more physical consistency, and physics interpretability. We summarize the potential implementations of physics-inspired generative models for SAR, as well as introducing some related work in other filed that would be serve as inspiration to us. 

\subsubsection{Physics-inspired Input}

Synthesizing data from physics-based models or simulators is a common way to harness domain knowledge. The synthesized data can be utilized either in in combination with real data to collaboratively train the generative model, or for independent pre-training. It is widely applied for SAR image despeckling where the training image pairs are obtained from simulating the speckle noise and injecting it into the high quality optical image data \cite{fracastoroDeepLearningMethods2021}.

Additionally, some studies proposed to extract the explainable features as the input of the generative model. For example, Zhang et al. \cite{zhangComparativeAnalysisEdge2021} proposed a SAR-to-Optical translation method based on paired features. For dual-polarized SAR images, the texture and edge features obtained from gray level concurrence matrix (GLCM) and canny operator of VV and VH channels are exploited. The pixel spectral features from multiple bands of the optical remote sensing images are also extracted, including near infrared and normalized difference vegetation index (NDVI).

\subsubsection{Physics-inspired Regularization}

In generative model training, the regularization term can be designed to constrain the outputs of the generative model which is derived from physical model or other scientific knowledge. It can be referred to some contents in Section \ref{sec:expprior}. In computer vision domain, the geometry, structure, depth, and edge information are extensively considered to constrain the inconsistency \cite{weiLiDeNeRFNeuralRadiance2024a,kimInfoNeRFRayEntropy2022,wu2024domain}.

A simple example for SAR is from literature \cite{marmanis2017artificial}, where a statistical regularization term is derived measuring the distance between gray-scale histogram of generated and real SAR images. In SAR image speckling applications, the statistical information was widely explored to constrain the model training. Vitale et al. \cite{vitaleMultiObjectiveCNNBasedAlgorithm2021a} proposed a regularization term that took into account the statistical properties of the speckle noise in the SAR image. The speckle is theoretically derived using the Rayleigh distribution with a predefined parameter. The proposed regularization term quantified the Kullback-Leibler divergence between the probability distribution function (pdf) of simulated speckle and the estimated ratio image, which is the ratio between the SAR image and the generated despeckled image. The objective is to impose a constraint on the network that requires it to generate the ratio image in accordance with the statistical properties of the speckle.

In addition to the statistical prior of SAR image, the physical parameters and scattering mechanisms of SAR were also explored in some researches. For example, the cross-entropy function was declared with the quantified physical parameter derived from polarimetric covariance matrix in \cite{songRadarImageColorization2018,dengQuadPolSARData2023a}. For surface scattering learning of SAR target, Wei et al. \cite{wei2024learning} proposed a regularization term to smooth the spatial optimization, that inhibited the occurrence of anomalies and abrupt transitions in the optimization parameters. 

\begin{figure*}[!htbp]
    \centering
    \includegraphics[width=0.8\textwidth]{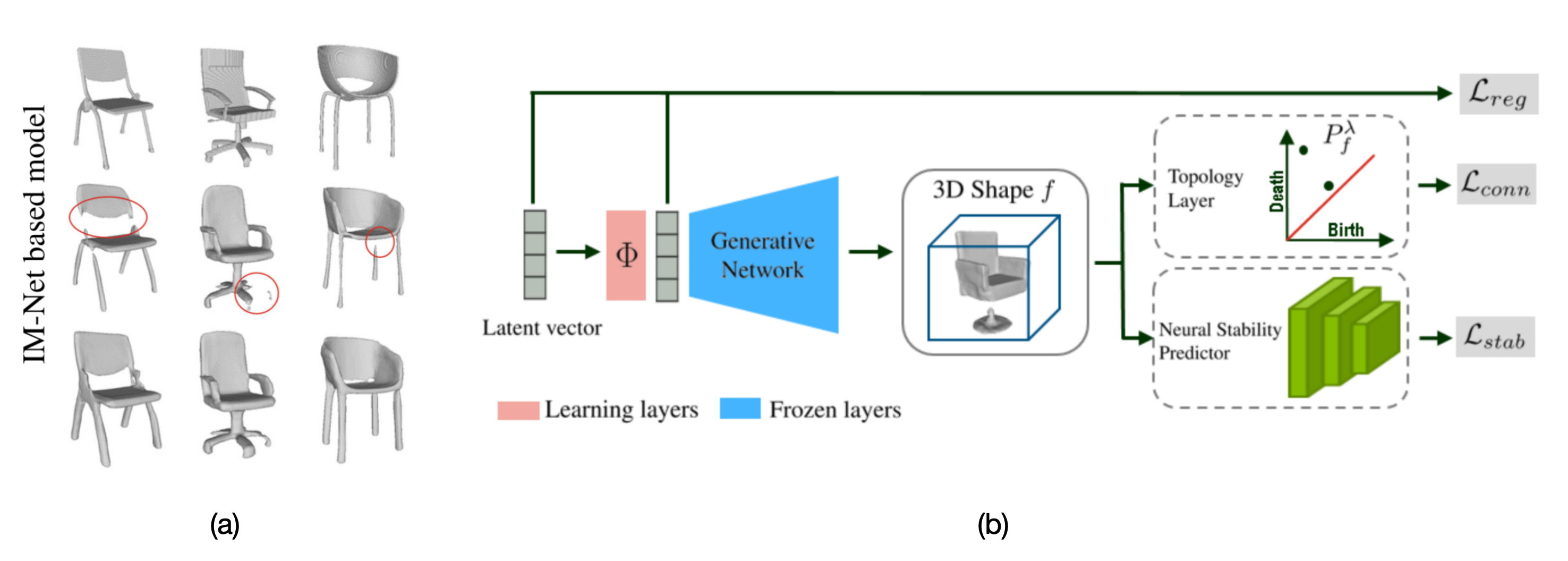}
    \caption{Mezghanni et al.  \cite{mezghanniPhysicallyawareGenerativeNetwork2021} proposed a physically-aware generative network for 3D shape reconstruction to address the problem of physics inconsistency. (a) Top: groundtruth, Center: without physical constraint, Bottom: with physical constraint. (b) Two regularization terms are designed to ensure the connectivity and stability subjected to gravity.}
    \label{fig:shapereg}
\end{figure*}

The geometry information is a critical physical constraint for the intended 3D reconstruction. Measuring the distance between the predicted 3D shape and the groundtruth is the conventional approach. Nevertheless, the vast solution space of this extreme non-convex problem presents significant optimization challenges. Moreover, the predicted shape may violate the physics knowledge of electromagnetic model or other disciplines. In computer vision domain, Mezghanni et al.  \cite{mezghanniPhysicallyawareGenerativeNetwork2021} proposed a physically-aware generative network for 3D shape reconstruction to address the problem of physics inconsistency, as shown in Fig. \ref{fig:shapereg}. The fundamental contribution is the designed physics-aware loss function in which the connectivity and physical stability of the generated shapes were promoted. In this framework, a neural network based stability predictor was designed and pre-trained with simulated data for enforcing the physical stability constraints.




\subsubsection{Physics-inspired Model Design}

Domain knowledge can also be integrated through a customized design of generative model architectures. One of them is substituting a part of the deep generative model with the well-established physical model. An interactive deep reinforcement learning (DRL) framework for angle prediction of SAR image was proposed in \cite{wang2024reinforcement} where DSR acted as the environment that provided the difference between the simulated SAR image with the current estimated angle and the measured data.

Another type of physics-inspired model design aims to modify the original architecture of the deep learning model based on the discipline of SAR physical process, such as the SAR image formation model. Lei et al. \cite{lei2024sar} recently proposed a SAR-NeRF model based on the neural radiance field, where the imaging geometry and volume rendering model for optical images is not applicable to SAR. To address this issue, the authors re-formulate the rendering process based on the mapping and projection principle for SAR image. Instead of using color and density to describe each pixel in natural image obtained from camera, the attenuation coefficients and scattering intensities in the 3D imaging space were derived analytically to implicitly represent the rendering equation. There are only a few studies introducing NeRF to SAR/ISAR image generation and reconstruction task in the current literature \cite{snyder2023extending,zhangCircularSARIncoherent2023,deng2024isar,liu2023ranerf,ehret2024radar}. The main challenge lies in the the different imaging mechanism of SAR data. We also reviewed some related work addressing the NeRF application on LiDAR which consider the specific imaging mechanism of the sensor instead of visual camera, or on satellite imagery with different imaging geometry. The traditional approach to synthesizing novel LiDAR views suffers from discretization artifacts and errors due to its idealized ray model, which doesn't account for LiDAR beam divergence and secondary returns. To this end, Huang et al. \cite{huang2023neural} proposed a Neural LiDAR Fields (NLF) that extended NeRF by incorporating a detailed model of the LiDAR sensing process, where the volume rendering is adapted to allow for more accurate representation of how LiDAR sensors measure distances and capture returns from emitted rays. Mari et al. \cite{mariSatNeRFLearningMultiView2022} proposed a Sat-NeRF model to address the issues of radiometric inconsistencies, shadows, and transient objects. Specifically, the RPC model of each input image is directly used to cast rays in the object space, enhancing the independence and accuracy compared to traditional pinhole camera models. Besides, Borts et al. \cite{bortsRadarFieldsFrequencySpace2024} proposed the Radar Fields of Frequency-Space Neural Scene Representations for FMCW Radar. Huang et al. \cite{huangDARTImplicitDoppler} proposed a NeRF-inspired method which uses radar-specific physics to create a reflectance and transmittance-based rendering pipeline for range-Doppler images. The above literature tackle the similar challenge of extending the current NeRF model to other imaging system which can serve as references.


\section{Datasets, Experiments, and Evaluation}

In this section, we first provide an overview of the open-source datasets used for SAR image generation task, and then summarize the existing evaluation techniques for generated SAR image. The evaluation methods include some quantitative metrics of image quality assessment (IQA), the assessments of the practical value of generated data for downstream tasks, and evaluations of the fidelity to the physical properties of SAR. In the experiments part, we provide several baseline models for SAR target generation on MSTAR benchmark dataset. The different evaluation metrics are also compared according to the generated results.


\begin{figure*}[!hbp]
    \centering
    \includegraphics[width=0.85\textwidth]{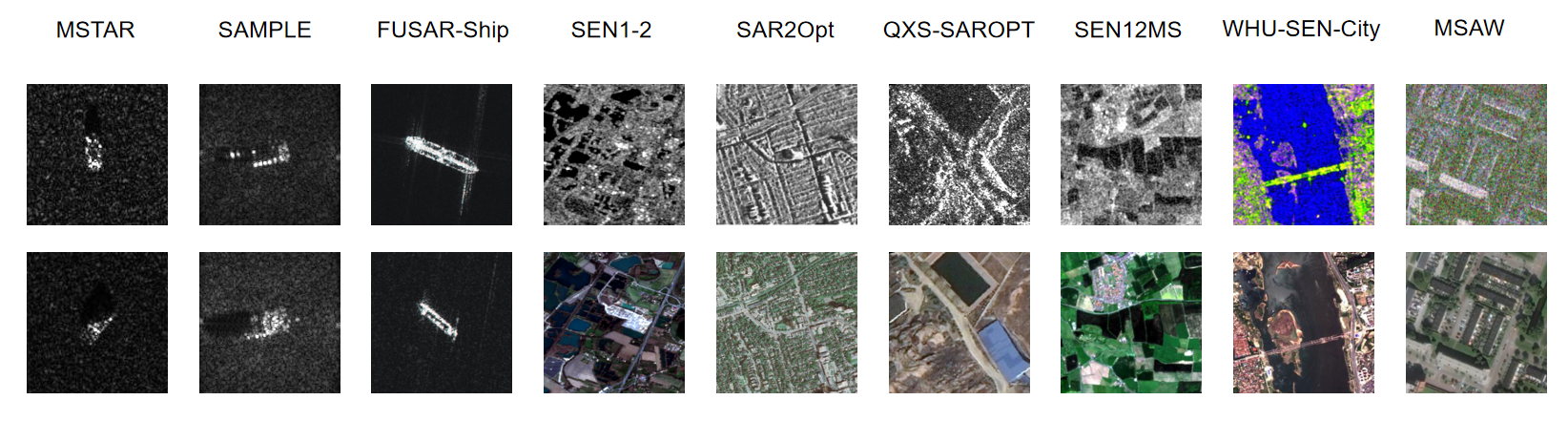}
    \caption{Visualization samples of datasets. For datasets used for SAR-to-Optical translation, the SAR images are on the top row while the paired optical images are on the bottom.}
    \label{fig:dataset}
\end{figure*}

\subsection{Datasets}

The common-used datasets are introduced as follows, with a glance in Fig. \ref{fig:dataset}.

\subsubsection{MSTAR} The moving and stationary target acquisition and recognition (MSTAR) dataset\cite{mstar} is the most widely used dataset in the sar target generation task. MSTAR consists of ten different types of vehicles collected at different viewing angles, at a resolution of 0.3m×0.3m, taken by X-band SAR sensor.The azimuth angle ranges from 0° to 360°, with an approximate interval of 1° to 2°. Typically, in the target generation task, the training set consists of images with a depression angle of 17°, while the images with a depression angle of 15° are used for the test. Several studies, including  \cite{caoLDGANSyntheticAperture2020,ohPeaceGANGANBasedMultiTask2021,huFeatureLearningSAR2021,songLearningGenerateSAR2022,guoCausalAdversarialAutoencoder2023} et al. all utilised MSTAR as the benchmark dataset to assess the generative performance of their models.

\subsubsection{FUSAR-Ship} FUSAR-Ship \cite{houFUSARShipBuildingHighresolution2020} is another dataset employed for the task of generating SAR target images, consisting of 15 main ship categories, 98 sub-categories, and many non-ship targets from Gaofen-3 SAR images. \cite{songLearningGenerateSAR2022,guoCausalAdversarialAutoencoder2023} both conducted additional assessments on the generative models' capacity to generate other real-world targets based on some typical typical ships in FUSAR-Ship.

\subsubsection{SAMPLE} The Synthetic and Measured Paired and Labeled Experiment (SAMPLE) dataset \cite{lewis2019sar} consists of SAR data from the MSTAR dataset and well-matched EM-based synthetic data whose imaging parameters are the same as those of MSTAR dataset. The azimuth angles of the SAMPLE dataset range from 10° to 80° with the depression angle in the range of 15° to 17°

\subsubsection{SEN1-2} SEN1-2 \cite{schmitt2018sen1} is perhaps the most commonly used public dataset for SAR-to-Optical generation task.The dataset contains 282384 pairs of SAR images acquired by Sentinel-1 at a maximum spatial resolution of 5 m and optical images obtained from Sentinel-2 satellites. The whole dataset including various landforms is made up of four subdatasets (SEN1-2-Spring, SEN1-2-Summer, SEN1-2-Fall, and SEN1-2-Winter) over all four seasons. The images size used in most works is usually set to 256 × 256 pixels. The dataset was employed in works such as \cite{zhaoComparativeAnalysisGANBased2022,liSARtoOpticalImageTranslation2020,leeCFCASETCoarsetoFineContextAware2023,weiCFRWDGANSARtoOpticalImage2023,baiConditionalDiffusionSAR2024,guoEdgePreservingConvolutionalGenerative2021,yangFGGANFineGrainedGenerative2022,liMultiscaleGenerativeAdversarial2022,fuentesreyesSARtoOpticalImageTranslation2019,yangSARtoopticalImageTranslation2022,zhangSARtoOpticalImageTranslation2024,wangSARtoOpticalImageTranslation2022} to evaluate the effectiveness of the SAR-to-Optical translation methods.

\subsubsection{SAR2Opt} The SAR2Opt dataset introduced in \cite{zhaoComparativeAnalysisGANBased2022} is specifically built for SAR-to-Optical image translation.The SAR images in the SAR2Opt datasetwere acquired at a spatial resolution of 1 m by TerraSAR-X in ten cities located in Asia, North America, Oceania, and Europe between 2007 and 2013. By manually choosing control locations, optical pictures were gathered from Google Earth Engine and paired with SAR images.
Coregistered SAR-to-optical image pairings were processed to obtain image patches of size 600 × 600 pixels. In total, 2,077 paired large-sized, high-resolution SAR-optical image patches are included in SAR2Opt.\cite{shiBraininspiredApproachSARtooptical2024,zhaoComparativeAnalysisGANBased2022} utilized the dataset for model performance evaluation.

\subsubsection{QXS-SAROPT} QXS-SAROPT dataset \cite{huang2021qxs} is also often used for SAR-to-optical image translation. QXS-SAROPT contains 20,000 SAR-optical patch-pairs from multiple scenes which covers three port cities: Qingdao, Shanghai, and San Diego at a high resolution of 1 m. The SAR and optical images are acquired from Gaofen-3 SAR satellite and Google Earth respectively. For convenient use in the translation task, the SAR-optical image pairs in QXS-SAROPT are cropped into small patches of 256 × 256 pixels. \cite{weiCFRWDGANSARtoOpticalImage2023,wangSARtoOpticalImageTranslation2022} both used QXS-SAROPT to prove the effectiveness of their methods.

\subsubsection{SEN12MS}The SEN12MS dataset \cite{schmitt2019sen12ms} can also support the SAR-to-optical image translation task. A total of 180,662 patch triplets consisting of corresponding Sentinel-1 dual-pol SAR images, Sentinel-2 multispectral images and and MODIS land cover maps comprises the whole SEN12MS dataset. The image patches set to 256 × 256 pixels in size are dispersed throughout all four of the planet's meteorological seasons and over all geographical masses. The SEN12MS dataset was used in \cite{yangFGGANFineGrainedGenerative2022} for translation.

\subsubsection{WHU-SEN-City} In \cite{wangSARtoOpticalImageTranslation2019a}, another dataset called WHU-SEN-City dataset was introduced for SAR-to-optical translation. The dataset contains 18542 paired training samples and 4566 paired test samples in all covering 32 Chinese cities. Two bands, VH and VV, having a spatial resolution of 20 m in the range direction and 22 m in the azimuth direction, make up the SAR images from Sentinel-1 and the paired optical images are from red, green, and blue bands of the Sentinel-2. \cite{yangFGGANFineGrainedGenerative2022, wangSARtoOpticalImageTranslation2019a} utilized the dataset for a test of translation.

\subsubsection{MSAW} 3401 groups of multisensor images, including SAR and optical images, are included in the Multi-Sensor All Weather Mapping (MSAW) Dataset \cite{shermeyerSpaceNetMultiSensorAll2020}. These images were supplied by Capella Space in collaboration with the Maxar Worldview-2 satellite and MetaSensing, respectively. Thus, MSAW is naturally suitable for SAR-to-optical translation, with images centrally cropped into a size of 512 × 512. In \cite{wangSARtoOpticalImageTranslation2022}, the authors used this dataset for translation experiment.

\subsection{General IQA Methods}

Image quality assessment (IQA) methods mainly aim to assign a score to image that can evaluate its quality from a visual perspective. Regarding evaluating the generated SAR images, the objective IQA metrics are widely applied. It can be categorized into full-reference (FR), no-reference (NR) and reduced-reference (RR) approaches \cite{wang2006modern} based on the availability of reference images. Another IQA metrics are designed specifically for SAR system, such as radiometric resolution, equivalent number of looks (ENL), etc. We investigated current literature concerning generated SAR image quality evaluation and summarized the involved IQA methods in Table \ref{tab:IQA}.

\begin{table*}[h!]
    \caption{IQA metrics used in SAR-related generation tasks}
    \label{tab:IQA}
    \centering
    \begin{tabular}{ccccccccccc}
    \toprule
    \textbf{IQA Metrics} & \textbf{PSNR}& \textbf{MSE} & \textbf{RMSE} & \textbf{SSIM} & \textbf{M-SSIM} & \textbf{CW-SSIM} & \textbf{VIF} & \textbf{FSIM} & \textbf{FSIMc} & \textbf{MGSM} \\
    \midrule
    \makecell[c]{\textbf{SAR Target} \\ \textbf{Generation}} & \cite{lei2024sar}&\cite{wangSARTargetImage2022}& \cite{giry2022sar}& \cite{sunAttributeGuidedGenerativeAdversarial2023, wangSARTargetImage2022} & \cite{zengATGANSARTarget2024} & - & - & \cite{zengATGANSARTarget2024} & - &\cite{daiCVGANCrossViewSAR2023, songLearningGenerateSAR2022}                                              \\
\makecell[c]{\textbf{S2O}} & \makecell[c]{\cite{shiBraininspiredApproachSARtooptical2024,  zhangComparativeAnalysisEdge2021, yangSARtoopticalImageTranslation2022}\\ \cite{leeCFCASETCoarsetoFineContextAware2023, zhangSARtoOpticalImageTranslation2024,    weiCFRWDGANSARtoOpticalImage2023}\\
\cite{doiGANBasedSARtoOpticalImage2020, guoSar2colorLearningImaging2022}\\ \cite{shiSARtoOpticalImageTranslating2022, wangSARtoOpticalImageTranslation2019a}\\
\cite{hwangSARtoOpticalImageTranslation2020, wangHybridCGANCoupling2022}}&
\makecell[c]{\cite{yangSARtoopticalImageTranslation2022, niuImageTranslationHighresolution2021}\\
\cite{guoEdgePreservingConvolutionalGenerative2021,zhangFeatureGuidedSARtoOpticalImage2020}\\
\cite{ guoSar2colorLearningImaging2022, wangHybridCGANCoupling2022}} & \makecell[c]{\cite{weiCFRWDGANSARtoOpticalImage2023, noaturnesAtrousCGANSAR2022}\\
\cite{doiGANBasedSARtoOpticalImage2020}} & \makecell[c]{\cite{baiConditionalDiffusionSAR2024, shiBraininspiredApproachSARtooptical2024, zhangComparativeAnalysisEdge2021,yangSARtoopticalImageTranslation2022, niuImageTranslationHighresolution2021, leeCFCASETCoarsetoFineContextAware2023}\\ 
\cite{doiGANBasedSARtoOpticalImage2020, guoSar2colorLearningImaging2022}\\
\cite{shiSARtoOpticalImageTranslating2022, wangSARtoOpticalImageTranslation2019a}\\
\cite{hwangSARtoOpticalImageTranslation2020, wangHybridCGANCoupling2022}} 
&  -        & \cite{shiBraininspiredApproachSARtooptical2024} & \cite{wangHybridCGANCoupling2022} & \makecell[c]{\cite{zhangFeatureGuidedSARtoOpticalImage2020}\\
\cite{wangHybridCGANCoupling2022}} & \makecell[c]{\cite{zuoSARtoOpticalImageTranslation2021}\\
\cite{wangSARtoOpticalImageTranslation2019a}} & - \\
\textbf{Despeckling} & \makecell[c]{\cite{lattariCycleSARSARImage2023, pereraSARDespecklingUsing2023, wangSARImageDespeckling2022, koSARImageDespeckling2022, hu2024sar}}  &  - & -  & \makecell[c]{\cite{lattariCycleSARSARImage2023, pereraSARDespecklingUsing2023, wangSARImageDespeckling2022, koSARImageDespeckling2022, hu2024sar}}& - & -  & -  & - & -  & -   \\

\midrule
\textbf{IQA Metrics} & \textbf{NGSC} & \textbf{UQI}  & \textbf{FID}  & \textbf{KID}  & \textbf{LPIPS} & \textbf{NIQE} & \textbf{QNR} & \textbf{OBS}  & \textbf{COSS}      & \textbf{NCC} \\
\midrule

\makecell[c]{\textbf{SAR Target} \\ \textbf{Generation}}      & \makecell[c]{\cite{daiCVGANCrossViewSAR2023, songLearningGenerateSAR2022}}  & - & \cite{zengATGANSARTarget2024} & - & \cite{lei2024sar}  & - & - & - & \cite{sunAttributeGuidedGenerativeAdversarial2023}                                                                         & \makecell[c]{\cite{sunAttributeGuidedGenerativeAdversarial2023, daiCVGANCrossViewSAR2023}\\ \cite{songLearningGenerateSAR2022}} \\
\textbf{S2O} & - & \cite{wangGeneratingHighQuality2018}         & \makecell[c]{\cite{baiConditionalDiffusionSAR2024, shiBraininspiredApproachSARtooptical2024}\\
\cite{wangSARtoOpticalImageTranslation2022, liMultiscaleGenerativeAdversarial2022} \\
\cite{fuReciprocalTranslationSAR2021,yangFGGANFineGrainedGenerative2022}\\
\cite{liDeepTranslationGAN2021, wangHybridCGANCoupling2022}\\
\cite{leeCFCASETCoarsetoFineContextAware2023}}& \makecell[c]{\cite{liDeepTranslationGAN2021, liMultiscaleGenerativeAdversarial2022}}  & \makecell[c]{\cite{leeCFCASETCoarsetoFineContextAware2023, weiCFRWDGANSARtoOpticalImage2023}\\
\cite{yangFGGANFineGrainedGenerative2022, guoSar2colorLearningImaging2022}} & \makecell[c]{\cite{yangSARtoopticalImageTranslation2022, yangFGGANFineGrainedGenerative2022}} & \cite{leeCFCASETCoarsetoFineContextAware2023} & \cite{songSARImageRepresentation2019}  & - &                   -                 \\
\midrule
\textbf{IQA Metrics} & \textbf{CHD} & \textbf{ERGAS} & \textbf{SAM}  & \textbf{Ds}    & \textbf{DG} & \textbf{ENL}  & \textbf{MoI}  & \textbf{MoR} & \textbf{M-index}      & \textbf{EPD-ROA}                                                                                                                                      \\
\midrule
\textbf{S2O} &         \cite{leeCFCASETCoarsetoFineContextAware2023} & \cite{leeCFCASETCoarsetoFineContextAware2023} &   \makecell[c]{\cite{leeCFCASETCoarsetoFineContextAware2023,wangSARtoOpticalImageTranslation2022, noaturnesAtrousCGANSAR2022}}  & \cite{leeCFCASETCoarsetoFineContextAware2023}   & \cite{wangGeneratingHighQuality2018}  & - & - & - & - & - \\
\textbf{Despeckling} & - &  - & - &  -  &  -  & \makecell[c]{\cite{wangSARImageDespeckling2022, lattariCycleSARSARImage2023, koSARImageDespeckling2022}} & \cite{koSARImageDespeckling2022}              & \cite{koSARImageDespeckling2022} & \cite{koSARImageDespeckling2022}  & \cite{koSARImageDespeckling2022}     \\     
\bottomrule
\end{tabular}
\end{table*}

Most literature applied the conventional FR-IQA metrics based on pixel statistical, structure similarity information, and mutual information, to describe the quality of simulated SAR image, for example, mean square error (MSE), peak signal-to-noise ratio (PSNR), structural similarity (SSIM), visual information fidelity (VIF) \cite{9348877, 9933645, wang2018generating}, as given in Table \ref{tab:IQA}. These methods assess the image quality based on human visual perception. For generated image with deep learning models, the Fréchet Inception Distance (FID) is also widely-used from the perspective of feature distribution \cite{heusel2017gans}. It basically applies the ImageNet pre-trained Inception network as the feature extractor and compare the distance of features. Due to the specific imaging mechanism of SAR, however, there may be lack of trustworthiness for evaluation \cite{wang2018synthetic,ju2023sargan,wang2023improved}. The NR models mainly rely on the image quality benchmark datasets that record the evaluation results of human beings for visual perception \cite{mittal2012no,mittal2012making,talebi2018nima,8576582}. Most SAR-to-Optical translation methods applied them for evaluation. For despeckling tasks, some specialized criteria for SAR image quality evaluation were widely used \cite{wangSARImageDespeckling2022, lattariCycleSARSARImage2023, koSARImageDespeckling2022}, such as edge-preservation index (EPI), despeckling gain measure (DG), equivalent number of look (ENL), etc.

\subsection{Assessment of Utility}

Since the main objective for most SAR image generation tasks is to improve the downstream model training, it is important to evaluate the quality of generated data according to their utilities as training data. Many advanced simulation approaches can synthesize SAR images with good visual quality \cite{lewis2019sar}, yet they have difficulties in real application \cite{song2021learning,liu2018sar}. Table \ref{tab:utility} summarize several assessment approaches in terms of utility. Most of them are conducted by training and testing a deep model with real or simulated SAR images and reporting the model performance for illustration. It basically contains training a model with real SAR images and testing with simulated ones \cite{ohPeaceGANGANBasedMultiTask2021,luanBCNetBackgroundConversion2024,yangSARtoopticalImageTranslation2022}, training a model with the mixed real and simulated SAR images and testing with real ones \cite{wangSARTargetImage2022,daiCVGANCrossViewSAR2023,sunAttributeGuidedGenerativeAdversarial2023,huFeatureLearningSAR2021,songLearningGenerateSAR2022,lei2024sar,zengATGANSARTarget2024}, training a model with simulated SAR images and testing with real ones \cite{ohPeaceGANGANBasedMultiTask2021,guoCausalAdversarialAutoencoder2023,sunAttributeGuidedGenerativeAdversarial2023}. We noticed that the above methods, in principle, are measuring the feature distribution discrepancy of real and generated data, which is the most significant obstacle that influences the data utility. However, they can only assess the utility of the global dataset instead of individuals as FR-IQA methods can achieve. Moreover, it is difficult to summarize a definite quantitative metric to illustrate the utility.

\begin{table*}[]
\centering
\caption{The summary of utility evaluation for SAR image generation.}
\label{tab:utility}
\begin{tabular}{cccc}
\toprule
\textbf{Utility Assessment}  &   \makecell[c]{\textbf{Train: Generated Data} \\ \textbf{Test: Real Data}}    & \makecell[c]{\textbf{Train: Mixed Data} \\ \textbf{Test: Real Data}}  & \makecell[c]{\textbf{Train: Real Data} \\ \textbf{Test: Generated Data}} \\
\midrule
\textbf{SAR target generation}      & \cite{ohPeaceGANGANBasedMultiTask2021, guoCausalAdversarialAutoencoder2023, sunAttributeGuidedGenerativeAdversarial2023} & \cite{wangSARTargetImage2022,daiCVGANCrossViewSAR2023,sunAttributeGuidedGenerativeAdversarial2023,huFeatureLearningSAR2021,songLearningGenerateSAR2022,lei2024sar,zengATGANSARTarget2024} & \cite{ohPeaceGANGANBasedMultiTask2021} \\ 
\textbf{Image Composition} & \cite{sunDSDetLightweightDensely2021, luanBCNetBackgroundConversion2024, kuangSemanticLayoutGuidedImageSynthesis2023}   & \cite{zhang2024ship} & \cite{luanBCNetBackgroundConversion2024} \\
\textbf{SAR-to-Optical translation} & - & - & \cite{yangSARtoopticalImageTranslation2022, fuentesreyesSARtoOpticalImageTranslation2019, sunSARTargetRecognition2022} \\
\bottomrule
\end{tabular}
\end{table*}

Guo et al. \cite{guoRecognitionRateSubstitution2022} proposed the Substitution Rate Curve (RSC) criteria based on the recognition rate changing for evaluation, where a machine learning model was trained multiply with different ratios of real and simulated data to record and aggregate the results. Similarly, the hybrid recognition rate curve (HRR) and the class-wise image quality assessment were proposed in literature \cite{yu2022new}, respectively. 

Besides evaluating the utility, some other work aim to improve the data utility of simulated SAR images with pre-processing, transfer learning, domain adaptation, or knowledge distillation \cite{han2024improving,sun2023gradual,he2021sar,zhang2022sar,10217035}. The image pre-processing methods, such as Gaussian noise \cite{choi2018despeckling,makitalo2010denoising ,gleich2009wavelet,inkawhich2021bridging}, filtering \cite{choi2018despeckling,gleich2009wavelet}, explicitly change the simulated data itself. However, these strategies are manually and empirically designed which are less flexible. The implicit methods mainly include transfer learning and domain adaptation, aiming at learning the representative knowledge from simulated data applicable to measured ones \cite{shi2022unsupervised}. Nevertheless, they cannot explain why the simulated SAR images are not qualified for utility clearly. From the user's perspective, it is crucial to be able to figure out the flaws of generated data at the level of human understanding, so as to improve the data utility and the simulation method. In \cite{zhuang2023sar,huang2024xfakejugglingutilityevaluation}, the authors proposed an evaluation and explanation framework to assess and improve the distribution inconsistency for the first time. A Bayesian deep neural network trained with real SAR images was considered as the probabilistic evaluator. The predicted uncertainty revealed the feature distribution distinction of the tested simulated data and the training real data. The explainer was constructed based on an auto-encoder where the latent code was optimized under the guidance of uncertainty, thus obtaining the counterfactual explanation of the input simulated data which demonstrated the inauthentic details.

Nevertheless, the utility assessment is still insufficient in research for SAR image generation. In face recognition field, there is a close research direction named face image quality assessment (FIQA) based on biometric utility, which can be considered as an important reference and promote the research development for SAR community. It also refers to learning-based FIQA, where the face image quality is correlated to the performance of a face recognition model. SER-FIQ \cite{ou2021sdd}, for example, as an unsupervised deep learning-based FIQA method, was proposed to evaluate the robustness of face embeddings directly with face utility. The face image was passed to several sub-networks of a modified FR network by using dropout. Images with high-utility are expected to possess similar face representations resulting in low variance.

\subsection{Assessment of Physical Reality}

The examination primarily focuses on determining whether the outcome of the GenAI model aligns with established SAR-related scientific understanding. The current exploited aspects include the consistency of polarimetric characteristics \cite{song2017radar}, sub-aperture decomposition properties \cite{mohammadiasiyabiComplexValuedEndtoEndDeep2023}, geometry information \cite{jiaSARImageGeneration2023}, surface parameter \cite{wei2024learning}, etc. 

When synthesizing a PolSAR image, for example, it is crucial to assess the polarimetric decomposition result applied to the generated PolSAR data. The mean absolute error (MAE), Bartlett distance, and coherency index of polarimetric covariance matrix were applied for evaluation in \cite{song2017radar}. The Cloude–Pottier entropy–$\alpha$ plane was introduced to verify the correctness of the reconstructed full-pol SAR images \cite{aghababaeiDeepLearningBasedPolarimetricData2023}.

For SAR target generation of multi-views, Jia et al. \cite{jiaSARImageGeneration2023} introduced the Mean Intersection over Union (mIoU) to evaluate the mesh reconstruction error, which demonstrated if the predicted SAR image was rendered in accordance with the ground truth geometry information. In \cite{wei2024learning}, the predicted surface parameters were verified with ground truth. In many applications, however, the ground truth of physical parameters may not available so that the assessment of physical reality cannot be realized.

\subsection{Experiments}

\begin{figure*}[!htbp]
    \centering
    \includegraphics[width=1.0\textwidth]{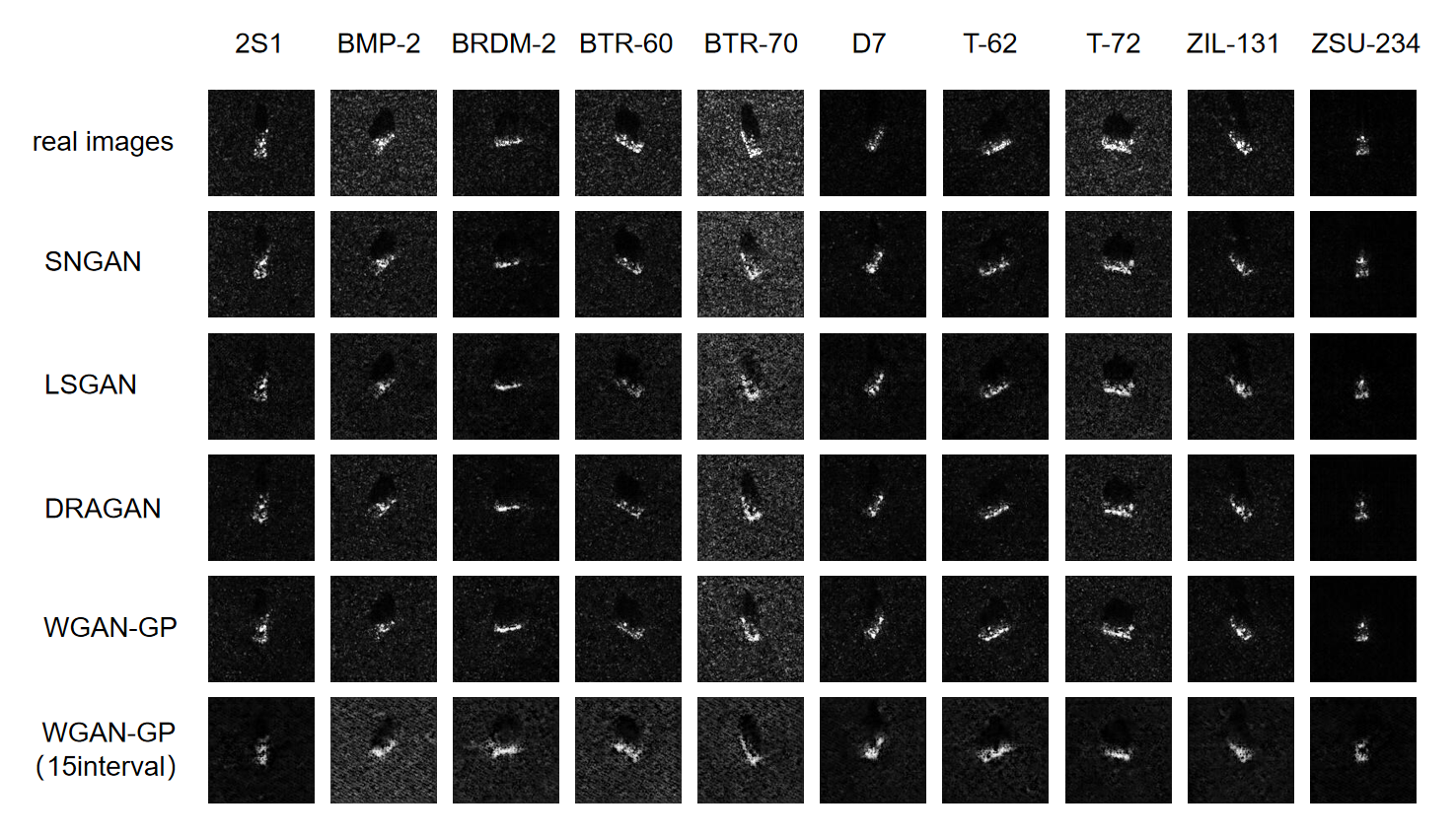}
    \caption{We implement some generative models, including SNGAN, LSGAN, DRAGAN, and WGAN-GP, and show the generated results. The models of row 2-5 are trained with all real samples covering the 360$^\circ$ azimuth angles with the interval of 1-2$^\circ$. As a comparison, the model of row 6 is trained with selected samples which have the azimuth angle interval of 15$^\circ$.}
    \label{fig:ex}
\end{figure*}

In this section, we present several baseline models for SAR target image generation and evaluate the generated data using various metrics. The source code and generated results are publicly accessible at \url{https://github.com/XAI4SAR/GenAIxSAR}. The baseline models are derived from ACGAN, utilizing class label $y$ and azimuth angle $\theta$ as conditional inputs. The angle information is encoded as $[\cos{\theta},\sin{\theta}]$. We experimented with four different GAN models, that are, SNGAN \cite{miyato2018spectral}, LSGAN \cite{mao2017least}, DRAGAN \cite{kodali2017convergence}, and WGAN-GP \cite{gulrajani2017improved}. Both the discriminator and generator were trained using the Adam optimizer with $\beta_1 = 0.5$ and $\beta_2 = 0.999$ and a learning rate of 0.0001. To stabilize the training process, each GAN model was trained with one discriminator iteration for every five generator iterations per epoch. The training data consisted of 10-class target images from the MSTAR dataset at a depression angle of 17$^\circ$ During the testing phase, label and angle information corresponding to a 15$^\circ$ depression angle were input to generate the target images. Some selected generated results are shown in Fig \ref{fig:ex}. By comparing the results of WGAN-GP trained with sufficient images and limited observation angles (15 $^\circ$ interval), it can be observed that the novel-view SAR target synthesis with a few observations is still challenging.

We conducted a series of experiments to evaluate the quality of the generated SAR images using various assessment methods, as presented in Table \ref{tab:acc}. The evaluation included four full-reference image quality assessment (IQA) metrics—PSNR, SSIM, FSIM, and VIF—as well as two no-reference IQA metrics—LPIPS and NIQE. In addition, we incorporated two utility evaluation methods: the recognition rate, which indicates the performance of a model trained on the generated SAR images when tested on real data, and Prob\_Eva, which measures the feature discrepancy between real and generated images by uncertainty of the probabilistic evaluator \cite{huang2024xfakejugglingutilityevaluation}. The best and worst results for each evaluation method are highlighted in red and blue, respectively.

The results indicate that most evaluation metrics favor the DRAGAN model, suggesting it produces images of the highest quality. Conversely, images generated by the LSGAN and WGAN-GP models tend to be of comparatively lower quality. However, it is noteworthy that in some instances, the evaluation results from SSIM and FSIM contradict those of other metrics. This observation underscores the potential unreliability of relying on a single evaluation metric for assessing the quality of generated SAR images, highlighting the importance of employing a diverse set of assessment methods to obtain a comprehensive evaluation from multiple perspectives.

\begin{table*}[]
\centering
\caption{The generated SAR images with SNGAN, LSGAN, DRAGAN, and WGAN-GP, are evaluated with different methods, including full-reference IQA metrics, no-reference IQA metrics, and utility evaluation methods. \textcolor{red}{Red} and \textcolor{blue}{Blue} denote the best and worst quality of each evaluation method, respectively.}
\label{tab:acc}
\begin{tabular}{ccccc}
\toprule
\multirow{2}{*}{\textbf{\makecell[c]{Evaluation Methods}}} & \multicolumn{4}{c}{\textbf{Generation Methods}} \\
\cmidrule(lr){2-5}
 & SNGAN & LSGAN   & DRAGAN  & WGAN-GP   \\
\midrule
\textit{Full-reference IQA metrics} \\
\textbf{PSNR} ($\uparrow$)      & 21.0708 & 21.2907 & \textcolor{red}{21.4791} & \textcolor{blue}{20.9877}  \\
\textbf{SSIM} ($\uparrow$)      & \textcolor{blue}{0.3266} & 0.3277 & \textcolor{red}{0.3403} & 0.3277  \\
\textbf{FSIM} ($\uparrow$)      & 0.7081 & \textcolor{red}{0.7260} & 0.7179 & \textcolor{blue}{0.7046}  \\
\textbf{VIF} ($\uparrow$)      & \textcolor{red}{0.0529} & 0.0509 & 0.0525 & \textcolor{blue}{0.0488}  \\
\midrule
\textit{No-reference IQA metrics} \\

\textbf{LPIPS} ($\downarrow$) \cite{zhang2018unreasonable} & 0.2466 & \textcolor{blue}{0.2571} & \textcolor{red}{0.2379} & 0.2496  \\
\textbf{NIQE} ($\downarrow$) \cite{6353522}      & 5.6628 & \textcolor{blue}{6.0181} & \textcolor{red}{5.3095} & 5.5613 \\  
\midrule
\textit{Utility Evaluation} \\
\textbf{Recognition Rate} ($\uparrow$)     & \textcolor{red}{0.8592} & 0.8445 & 0.8539 & \textcolor{blue}{0.8440}  \\
\textbf{Prob\_Eva ($\downarrow$)} \cite{huang2024xfakejugglingutilityevaluation}   & 0.1244 & \textcolor{blue}{0.1351} & \textcolor{red}{0.1222} & 0.1335  \\

\bottomrule
\end{tabular}
\end{table*}







\begin{figure*}[!htbp]
    \centering
    \includegraphics[width=0.8\textwidth]{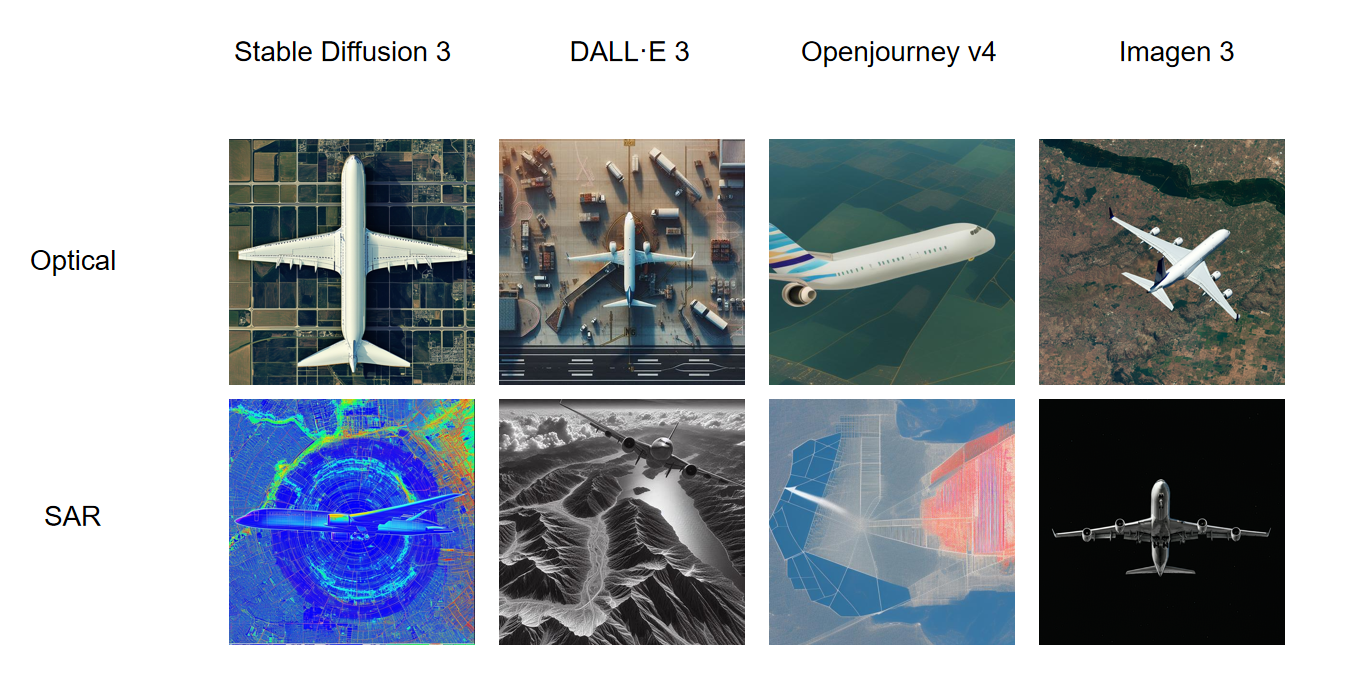}
    \caption{The generation results of text-to-image generative model. The text prompts are 'a remote sensing image whose content is an airplane' and 'a Synthetic Aperture Radar image whose content is an airplane' respectively. The models almost have the ability to generate remote sensing images but can hardly understand what SAR images are.}
    \label{fig:t2ir}
\end{figure*}

\section{Challenges and Perspectives}

This section presents the challenges and perspectives of SAR image generation and evaluation. Experimental data will be presented for analysis, demonstrating the shortcomings of the current methodologies. This discussion will focus on the future trends in achieving trustworthy SAR image generation and evaluation.

\subsection{Large-scale Datasets with Multimodal Information}


SAR image generation, compared with vision domain, is still in the early stage. The multi-view target generation task, for example, requires SAR target images obtained from various observation angles. The current researches mostly applied MSTAR or SAMPLE datasets to conduct experiments, but the data volume is limited and the target categories are behind the times. The generalization ability cannot be evaluated thoroughly with the several limited vehicle types. Some multimodal image synthesis applications correlated to SAR, such as text-to-image, layout-to-image, scene graph-to-image generation, have rarely been explored in SAR community. The primary obstacle is the deficiency of large-scale datasets, especially for multi-modality. The current cutting-edge generative AI models, such as Stable Diffusion, are constructed based on vision-language foundation model, trained with a very large number of multimodal data. There are two possible solutions to this challenge. The first is to integrate the current SAR image datasets, including SAR target detection and recognition, optical and SAR image fusion, and re-organize them with the assistance of other vision-language foundation models for various SAR image generation applications. The second is to propose more comprehensive SAR image datasets with multi-source information and knowledge, including 3D model, EM simulation, measured SAR images, etc.

\subsection{Abreast of GenAI Technology}

Most researches for SAR image generation still focus on GAN-based methods, but some cutting-edge GenAI-related technologies are still to be developed in this filed. The large-scale text-to-image generation diffusion model and the efficient 3D Gaussian splatting, for example, have never been explored in SAR domain so far. Indeed, they demonstrate obvious difficulties for SAR image applications due to the very different imaging geometry and electromagnetic characteristics of SAR. The preliminary attempts are necessary to explore the potentials and crucial scientific problems of them. 

\subsection{Continued Development Based on Text-to-Image Generative Model}

Current text-to-image generative foundation models have achieved remarkable performance in various domains, but their application to SAR data presents significant challenges. We experimented with several existing generative models, including Stable Diffusion 3, DALL-E 3, Openjourney v4, and Imagen 3, to generate optical remote sensing images and SAR images using similar text prompts. The results, shown in Fig. \ref{fig:t2ir}, indicate that these models can produce optical remote sensing images to varying degrees when provided with appropriate text prompts. However, they struggle significantly with understanding and generating "synthetic aperture radar" images. Training a text-to-image generative foundation model from scratch typically requires billions of paired images and text descriptions, which is particularly challenging in the SAR domain due to the scarcity of such data. A potential solution is to adapt existing text-to-image generative models for SAR applications through continued research, similar to the approach taken by ControlNet. This could be achieved using parameter-efficient fine-tuning methods like Low-Rank Adaptation (LoRA). Another approach is to decompose the text-to-SAR image generation task into two steps: first, generating the optical image from the text and then translating the optical image into a SAR image. However, as noted in previous research, the optical-to-SAR translation is particularly challenging due to its ill-posed nature, which is influenced by geometry and other imaging parameters. Therefore, to achieve more precise control, the text prompts should consider sensor parameters.

\subsection{Integration of Knowledge and AI}

The scientific understanding of SAR is often underestimated in much of the research reviewed in this survey. Hybrid modeling approaches, including AI-empowered physical modeling and physics-inspired generative AI (GenAI), remain in their infancy. Only a small portion of the literature has delved deeply into this research direction, as summarized in Section V. Key aspects such as scattering center properties, clutter distributions, and electromagnetic (EM) simulations have rarely been explored or integrated with generative AI models. In the future, greater emphasis should be placed on the synergy between AI and domain knowledge. For SAR image simulation and generation, leveraging AI to enhance the efficiency and flexibility of traditional physical models, while incorporating domain-specific knowledge to improve the interpretability and reliability of GenAI-generated results, are critical avenues for exploration. This involves not only developing physics-inspired regularization methods but also constructing more trustworthy GenAI architectures that are cognizant of SAR's physical principles. Considering the intrinsic interdisciplinary characteristics, we should obtain more inspirations from other related realms, as we discussed in Section V.

\subsection{Synergizing Generation and Perception}

The objective of most SAR image generation is to tackle the data hungry issue in specific application scenarios. The generated image quality should meet the satisfactory of the SAR image interpretation model. Evaluating them only with IQA metrics based on visual perception is unilateral. Many advanced simulation approaches can synthesize SAR images with good visual quality, yet they have difficulties in real application. It should be noted that the inconsistent distribution between generated and real data would be one of the most significant issues that influences the utility of the generated SAR image. As a result, evaluating the generated SAR images in terms of physical reality and data utility should be paid more attention in the future. Moreover, synergizing the generation and perception in a unified framework is necessary. To this end, the generative model can be additionally optimized in accordance with the utility of them in the application task.

\bibliographystyle{IEEEtran}
\bibliography{IEEEabrv,ref}












\section{Biography Section}
 



\begin{IEEEbiographynophoto}{Zhongling Huang}(Member, IEEE) (Northwestern Polytechnical University, huangzhongling@nwpu.edu.cn) received the B.Sc. degree from Beijing Normal University in 2015, and the Ph.D. degree from the University of Chinese Academy of Sciences (UCAS) and the Aerospace Information Research Institute, Chinese Academy of Sciences, in 2020. She was a Visiting Scholar with the German Aerospace Center from 2018 to 2019. She is currently an associate professor at the BRain and Artificial INtelligence Laboratory (BRAIN LAB), School of Automation, Northwestern Polytechnical University, Xi’an, China. Her research interests include explainable deep learning for synthetic aperture radar (XAI4SAR), SAR image interpretation, deep learning, and remote sensing data mining.
\end{IEEEbiographynophoto}
\begin{IEEEbiographynophoto}{Xidan Zhang} (Northwestern Polytechnical University) received the B.Sc. degree from Northwestern Polytechnical University in 2023. He is currently a graduate student at the BRain and Artificial INtelligence Laboratory (BRAIN LAB), School of Automation, Northwestern Polytechnical University, Xi’an, China. His research interests include SAR image generation and deep learning.
\end{IEEEbiographynophoto}
\begin{IEEEbiographynophoto}{Zuqian Tang} (Northwestern Polytechnical University) received the B.Sc. degree from Harbin Engineering University in 2024. He is currently a graduate student at the BRain and Artificial INtelligence Laboratory (BRAIN LAB), School of Automation, Northwestern Polytechnical University, Xi’an, China. His research interests include SAR image generation and deep learning.
\end{IEEEbiographynophoto}
\begin{IEEEbiographynophoto}{Feng Xu}(Senior Member, IEEE) (Key Laboratory for Information Science of Electromagnetic Waves (MoE), Fudan University, fengxu@fudan.edu.cn) received the B.E. degree (Hons.) in information engineering from Southeast University, Nanjing, China, in 2003, and the Ph.D. degree (Hons.) in electronic engineering from Fudan University, Shanghai, China, in 2008.
From 2008 to 2010, he was a Post-Doctoral Fellow with the NOAA Center for Satellite Application and Research (STAR), College Park, MD, USA. From 2010 to 2013, he was a Research Scientist with Intelligent Automation Inc., Rockville, MD, USA. Since 2013, he has been a Professor with the School of Information Science and Technology. He is currently the Vice Dean of the School of Information Science and Technology and the Director of the Key Laboratory for Information Science of Electromagnetic Waves (MoE), Shanghai. His research interests include electromagnetic scattering theory, synthetic aperture radar (SAR) information retrieval, and intelligent radar systems.
Dr. Xu was a recipient of the Early Career Award of the IEEE Geoscience and Remote Sensing Society in 2014 and SUMMA Graduate Fellowship in the advanced electromagnetics area in 2007. Among other honors, he was awarded the Second-Class National Nature Science Award of China in 2011 and the First-Class Nature Science Award of Shanghai in 2022. He is the Founding Chair of the IEEE GRSS Shanghai Chapter and an AdCom member of IEEE GRSS. He also served as an Associate Editor for IEEE Geoscience and Remote Sensing Letters (2014–2020) and IEEE Transactions on Geoscience and Remote Sensing (2023).
\end{IEEEbiographynophoto}
\begin{IEEEbiographynophoto}{Mihai Datcu}(Fellow, IEEE) (National Science and Technology University POLITEHNICA Bucharest, mihai.datcu@upb.ro) received the M.S. and Ph.D. degrees in electronics and telecommunications from the University Politehnica of Bucharest (UPB), Bucharest, Romania, in 1978 and 1986, respectively, and the habilitation a Diriger Des Recherches degree in computer science from the University Louis Pasteur, Strasbourg, France, in 1999. His research interests include information theory, signal processing, explainable and physics-aware artificial intelligence, computational imaging, and quantum machine learning with applications in EO. He is a member of the ESA Working Group Big Data from Space. He was a recipient of the Romanian Academy Prize Traian Vuia in 1987, the best paper award and the IEEE Geoscience and Remote Sensing Society Prize in 2006, the National Order of Merit with the rank of Knight awarded by the President of Romania in 2008, the Chaire d’excellence international Blaise Pascal 2017, the 2018 Ad Astra Award for Excellence in Science, and the IEEE GRSS David Landgrebe Award in 2022. He has served as a Co-organizer for international conferences and workshops and as Guest Editor for a special issues on AI and Big Data of the IEEE and other journals.
\end{IEEEbiographynophoto}
\begin{IEEEbiographynophoto}{Junwei Han}(Fellow, IEEE) (Northwestern Polytechnical University, jhan@nwpu.edu.cn) received the B.S., M.S., and Ph.D. degrees in pattern recognition and intelligent systems from Northwestern Polytechnical University, Xi’an, China, in 1999, 2001, and 2003, respectively. He is currently a professor at the BRain and Artificial INtelligence Laboratory (BRAIN LAB), School of Automation, Northwestern Polytechnical University. His research interests include computer vision and brain-imaging analysis.
\end{IEEEbiographynophoto}

\vfill

\end{document}